\documentclass[10pt,twocolumn,letterpaper]{article}

\usepackage{iccv}
\usepackage{times}
\usepackage{epsfig}
\usepackage{graphicx}
\usepackage{amsmath}
\usepackage{amssymb}
\usepackage{mmstyles}
\usepackage{enumitem}
\usepackage{caption}
\usepackage{float}
\usepackage{booktabs}

\usepackage[breaklinks=true,bookmarks=false]{hyperref}

 \iccvfinalcopy %

\def\iccvPaperID{6095} %

\ificcvfinal\fi

\begin{document}

\title{\vspace{-0.8cm}Towards Vivid and Diverse Image Colorization with Generative Color Prior}
\author{
	Yanze Wu, \hspace{9pt}Xintao Wang, \hspace{9pt}Yu Li, \hspace{9pt}Honglun Zhang, \hspace{9pt}Xun Zhao, \hspace{9pt}Ying Shan \\
	\vspace{-0.05cm}
	{Applied Research Center (ARC), Tencent PCG} \\
	{\tt\small\{yanzewu,xintaowang,ianyli,honlanzhang,emmaxunzhao,yingsshan\}@tencent.com}\\
}

\twocolumn[{%
    \renewcommand\twocolumn[1][]{#1}%
     \vspace{-0.5cm}
    \maketitle
    \begin{center}
        \centering
        \vspace{-0.7cm}
        \includegraphics[width=\textwidth]{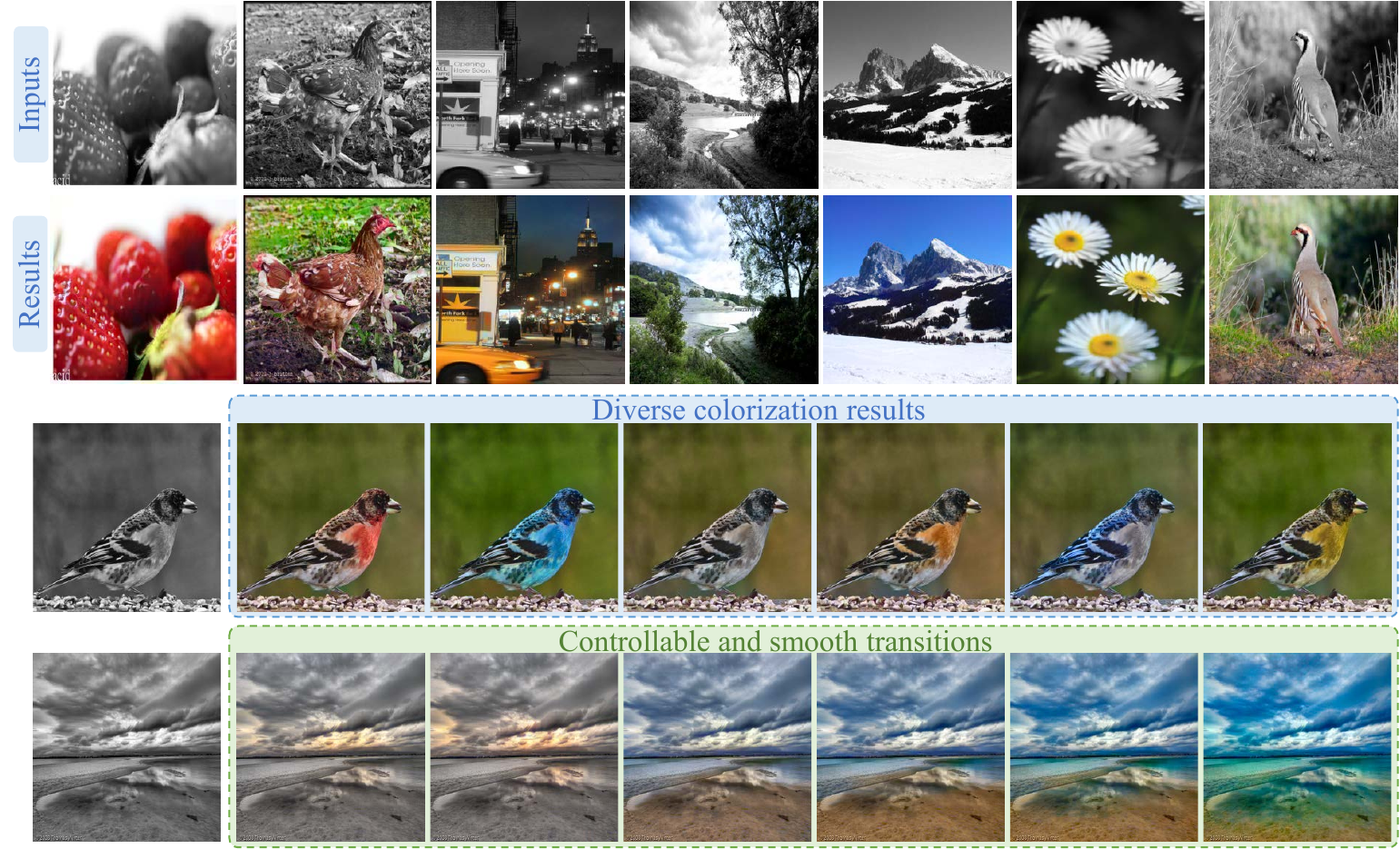}
        \vspace{-0.6cm}
        \captionof{figure}{By leveraging the generative color prior and the delicate designs, our method is capable of generating vivid and diverse colorization results without relying on external exemplars. Besides, our method could attain controllable and smooth transitions by walking through the GAN latent space.}
        \label{fig:teaser}
    \end{center}%
}]
\ificcvfinal\thispagestyle{empty}\fi

\begin{abstract}
Colorization has attracted increasing interest in recent years.
Classic reference-based methods usually rely on external color images for plausible results. A large image database or online search engine is inevitably required for retrieving such exemplars.
Recent deep-learning-based methods could automatically colorize images at a low cost.
However, unsatisfactory artifacts and incoherent colors are always accompanied.
In this work, we propose GCP-Colorization that leverages the rich and diverse color priors encapsulated in a pretrained Generative Adversarial Networks (GAN) for automatic colorization.
Specifically, we first ``retrieve'' matched features (similar to exemplars) via a GAN encoder and then incorporate these features into the colorization process with feature modulations. 
Thanks to the powerful generative color prior (GCP) and delicate designs, our GCP-Colorization could produce vivid colors with a single forward pass.
Moreover, it is highly convenient to obtain diverse results by modifying GAN latent codes. 
GCP-Colorization also inherits the merit of interpretable controls of GANs and could attain controllable and smooth transitions by walking through GAN latent space.
Extensive experiments and user studies demonstrate that GCP-Colorization achieves superior performance than previous works. Codes are available at \href{https://github.com/ToTheBeginning/GCP-Colorization}{https://github.com/ToTheBeginning/GCP-Colorization}.
\end{abstract}

\vspace{-0.4cm}
\section{Introduction}
Colorization, the task of restoring colors from black-and-white photos, has wide applications in various fields, such as photography technologies, advertising or film industry~\cite{Iizuka_2019,Vitoria_2020}.
As colorization requires estimating missing color channels from only one grayscale value, it is inherently an ill-posed problem.
Moreover, the plausible solutions of colorization are not unique (\eg, cars in black, blue, or red are all feasible results).
Due to the uncertainty and diversity nature of colorization, it remains a challenging task.

Classic reference-based methods (\eg~\cite{He_2018,Lu_2020,Xu_2020}) require additional example color images as guidance.
These methods attempt to match the relevant contents between exemplars and input gray images, and then transfer the color statistics from the reference to the gray one. 
The quality of generated colors (e.g., realness and vividness) is strongly dependent on the reference images.
However, retrieving desirable reference images requires significant user efforts.
A recommendation system for this can be a solution, but it is still challenging to design such a retrieval process. 
A large-scale color image database or online search engine is inevitably required in the system.

Recently, convolutional neural network (CNN) based colorization methods~\cite{Cheng_2015,Deshpande_2015,Isola_2017} have been proposed to facilitate the colorization task in an automatic fashion. 
These methods learn to discover the semantics~\cite{Larsson_2016,Zhang_2016} and then directly predict the colorization results.
Though remarkable achievements in visual qualities are achieved, unsatisfactory artifacts and incoherent colors are always accompanied.

In this work, we propose GCP-Colorization to 
exploit the merits of both reference-based and CNN-based methods, \ie, achieving realistic and vivid colorization results on par with reference-based methods, while keeping them automatic at a low cost. 
Inspired by recent success in Generative Adversarial Network (GAN)~\cite{brock2018large, karras2018stylegan}, our core idea is to leverage the most relevant image in the learned GAN distribution as the exemplar image.
The rich and diverse color information encapsulated in pretrained GAN models, \ie, generative color prior, allows us to
circumvent the explicit example retrieval step and integrate it into the colorization pipeline as a unified framework.
Specifically, we first `retrieve' matched features (similar to exemplars) via a GAN encoder and then incorporate these features into the colorization process with feature modulations. 
In addition, GCP-Colorization is capable of achieving diverse colorization from different samples in the GAN distribution or by modifying GAN latent codes. 
Thanks to the interpretable controls of GANs, GCP-Colorization could also attain controllable and smooth transitions by walking through GAN latent space.

We summarize the contributions as follows.
(1) We develop the GCP-Colorization framework to leverage rich and diverse generative color prior for automatic colorization.
(2) Our method allows us to obtain diverse outputs. Controllable and smooth transitions can be achieved by manipulating the code in GAN latent space.
(3) Experiments show that our method is capable of generating more vivid and diverse colorization results than previous works.

\section{Related Work}
\noindent \textbf{User assisted colorization.}
Early colorization methods are mostly interactive and require users to draw color strokes on the gray image to guide the colorization. 
The propagation from local hints to all pixels is usually formulated in a constrained optimization manner~\cite{Levin_2004} that tries to assign two pixels with the same color if they are adjacent and similar under similarity measures. 
Attempts in this direction focus on finding efficient propagation~\cite{Yatziv_2006} with different similarity metrics by hand-crafting~\cite{Qu_2006} or learning~\cite{Endo_2016}, building long-range relations\cite{Xu_2009}, or using edge detection to reduce bleeding~\cite{Huang_2005}. 
While these methods can generate appealing results with careful interactions, it demands intensive user efforts. 
The work of~\cite{Zhang_2017} alleviates this issue by using sparse color points, and introduces a neural network to colorize from these hints. 
Some methods also propose to use global hints like color palettes~\cite{Bahng_2018,Chang_2015} instead of dense color points as constraints.

\noindent \textbf{Reference-based methods} try to transfer the color statistics from the reference to the gray image using correspondences between the two based on low-level similarity measures~\cite{Liu_2008}, semantic features~\cite{Charpiat_2008}, or super-pixels~\cite{Chia_2011, Gupta_2012}. Recent works~\cite{He_2018,Lu_2020,Xu_2020} adopt deep neural network to improve the spatial correspondence and colorization results.
Though these methods obtain remarkable results when suitable references are available, the procedure of finding references is time-consuming and challenging for automatic retrieval system~\cite{Chia_2011}.

\noindent \textbf{Automatic colorization} methods are generally learning based~\cite{Cheng_2015,Deshpande_2015,Isola_2017,kumar2021colorization}. 
Pre-trained networks for classification are used for better semantic representation~\cite{Larsson_2016}.
Two branch dual-task structures are also proposed in~\cite{Iizuka_2016,Vitoria_2020,Zhao_2019,Zhao_2018} that jointly learn the pixel embedding and local (semantic maps) or global (class labels) information.
The recent work~\cite{Su_2020} investigates the instance-level features to model the appearance variations of objects.
To improve the diversity of generated colors, Zhang \etal~\cite{Zhang_2016} uses a class-rebalancing scheme, while some methods propose to predict per-pixel color distributions~\cite{Deshpande_2017,Messaoud_2018,Royer_2017} instead of a single color.
Yet, their results typically suffer from visual artifacts like unnatural and incoherent colors.

\begin{figure*}
	\begin{center}
		\vspace{-0.2in}
		\includegraphics[width=0.9\linewidth]{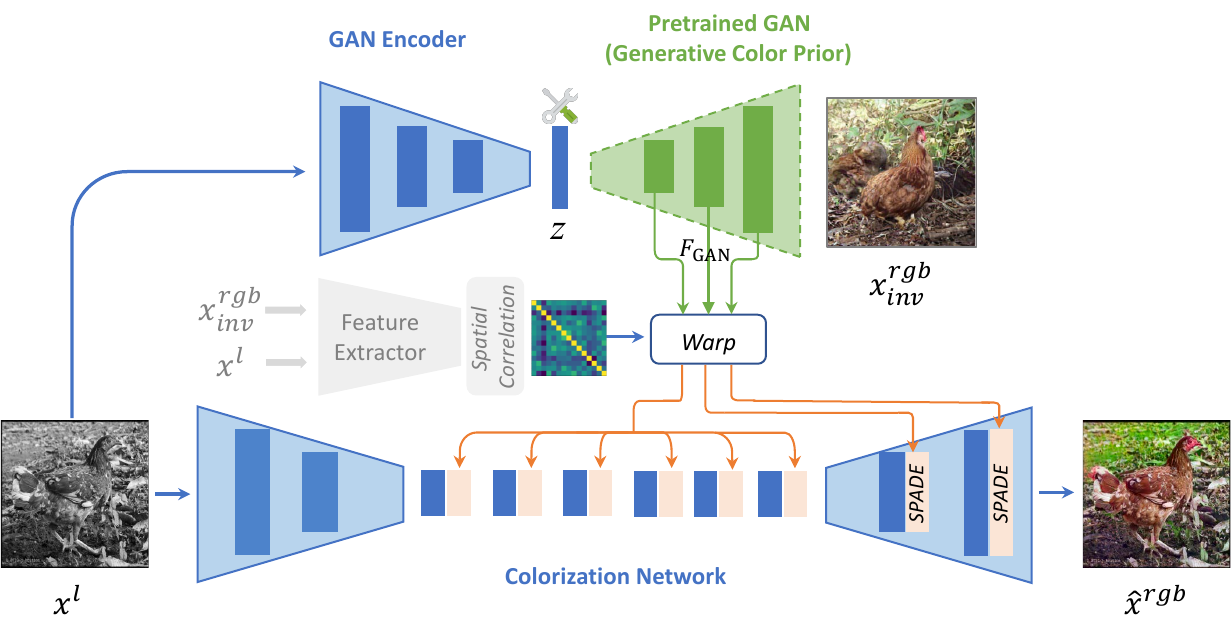}
	\end{center}
	\vspace{-0.7cm}
	\caption{\textbf{Overview of GCP-Colorization framework}. Given a grayscale image $x^l$ as input, our framework first produces the most relevant features $F_{\text{GAN}}$ and an inversion image $x_{inv}^{rgb}$ from a pretrained generative network as generative color priors. After that, we calculate a spatial correlation matrix from $x^l$ and $x_{inv}^{rgb}$, and warp GAN features $F_{\text{GAN}}$ for alignment. The warped features are used to modulate the colorization network by spatially-adaptive denormalization (SPADE) layers. Finally, a vivid colorized image can be generated with the colorization network. In addition, controllable and diverse colorization results, together with smooth transitions could be achieved by adjusting latent code $z$.}
	\label{fig:framework}
	\vspace{-0.4cm}
\end{figure*}

\noindent \textbf{Generative priors} of pretrained GANs~\cite{brock2018large,karras2018pggan,karras2018stylegan,karras2020stylegan2} is previously exploited by GAN inversion~\cite{abdal2019image2stylegan,gu2020mGANprior,menon2020pulse,pan2020dgp,zhu2020domain,zhu2016generative}, which aims to find the closest latent codes given an input image.
In colorization, they first `invert' the grayscale image back to a latent code of the pretrained GAN, and then conduct iterative optimization to reconstruct images.
Among them, mGANprior~\cite{gu2020mGANprior} attempts to optimize multiple codes to improve the reconstruction quality.
In order to reduce the gap between training and testing images, DGP~\cite{pan2020dgp} further jointly finetunes the generator and the latent codes.
However, these results struggle to faithfully retain the local details, as the low-dimension latent codes without spatial information are insufficient to guide the colorization. In contrast, our GCP-Colorization with spatial modulation enables prior incorporation on multi-resolution spatial features to achieve high texture faithfulness. 
In addition, our GCP-Colorization is feed-forward and does not require expensive iterative optimization for each instance.
Recent works on face restoration~\cite{wang2021gfpgan,Yang2021GPEN} also utilize generative priors for restoration and colorization. 

\section{Method}
\subsection{Overview}
With a grayscale image $x^l\in\mathbb{R}^{1\times H\times W}$ as input, the objective of colorization is to predict the two missing color channels $a$ and $b$ in the original color image $x^{lab}\in\mathbb{R}^{3\times H\times W}$, where $l$ and $a, b$ represent the luminance and chrominance in CIELAB color space, respectively.
In this work, we aim to leverage rich and diverse color priors encapsulated in a pretrained generative network to guide the colorization.
The overview of our GCP-Colorization framework is shown in Figure~\ref{fig:framework}. It mainly consists of a 
\textit{pretrained GAN} $\cG$,
a \textit{GAN Encoder} $\cE$, and a \textit{Colorization Network} $\cC$.

Given the grayscale image $x^l$, \textit{GAN Encoder} $\cE$ is responsible for mapping $x^l$ into a latent code $z$, which can serve as an input of $\cG$.
The \textit{pretrained GAN} $\cG$ then generates the most relevant color image (denoted as inversion image $x_{inv}^{rgb}$) to the grayscale input. In addition to using the inversion image directly, the prior features $F_{\text{GAN}}$ from intermediate layers are more informative and are leveraged to guide the colorization. However, these `retrieved' features usually do not spatially align  with the grayscale input. To solve this, we first perform an alignment step to warp $F_{\text{GAN}}$ based on the correspondence between $x^l$ and $x_{inv}^{rgb}$. After that, we use the aligned GAN features to modulate the multi-scale spatial features in \textit{Colorization Network} $\cC$ with effective SPADE (spatially-adaptive denormalization) layers~\cite{park2019SPADE}, producing the final colorized image $\hat{x}^{rgb}$.

\subsection{Generative Color Prior and GAN Encoder}
Recently, GAN has achieved tremendous advances in unsupervised image generation. Once trained on an adequate number of images, GAN can map arbitrary random noises into realistic and high-resolution images with impressive details and rich colors, which motivates us to employ pre-trained GAN models to assist the colorization task.
As illustrated above, we need to find a latent code $z$ so that $\cG$ can take it as the input and produce the GAN features $F_{\text{GAN}}$ from its intermediate layers which are most relevant to $x^l$. To achieve this goal, we constrain the generated image from code $z$ to have similar content with $x^l$ but contains rich chrominance information. The process of finding a certain $z$ corresponding to a target image is termed as \textit{GAN inversion}. A typical inversion way is to randomly initialize $z$ and optimize it to reduce the difference between the generated image and the target one. However, such a method requires expensive iterative optimization at inference and thus is not applicable for real-world scenarios. Instead, we choose to train a GAN encoder $\cE$ to map arbitrary grayscale images into latent codes, \ie, 
\begin{equation}
\label{eq:inv}
\begin{split}
&z = \cE(x^l),\\
&x_{inv}^{rgb}, F_{\text{GAN}} = \cG(z),
\end{split}
\end{equation}
where $\cG(z)$ simultaneously outputs the inversion image $x_{inv}^{rgb}$ and the intermediate GAN features $F_{\text{GAN}}$.

\paragraph{Discussion.} It is well known that the optimization-based inversion methods could generate more accurate inversion images than encoder-based inversion methods~\cite{zhu2020domain,zhu2016generative}. However, in our colorization scenarios, we do not need such precise and high-quality inversion images. A coarse inversion result already contains sufficient color information and the colorization network (illustrated in the next section) will further refine the corresponding inversion features. In addition, we use the GAN features as extra information to guide the colorization, which could provide 
complementary information to the inversion image.

\subsection{Colorization Network}
The colorization network $\cC$ consists of two downsampling blocks, six residual blocks and two upsampling blocks, which proves to be an effective architecture used in many image-to-image translation tasks~\cite{Wang_2018}. It takes the grayscale image $x^l$ as input and predicts two missing color channels $\hat{x}^{ab}$. A color image $\hat{x}^{lab}$ can then be obtained by concatenating the luminance channel $x^l$ and the chrominance channels $\hat{x}^{ab}$. 
As the two color spaces RGB and LAB can be converted to each other, $\hat{x}^{rgb}$ can also be obtained.

In order to use the prior color features $F_{\text{GAN}}$ to guide the colorization, and to better preserve the color information of $F_{\text{GAN}}$, we use spatially-adaptive denormalization (SPADE)~\cite{park2019SPADE} to modulate the colorization network. 
However, $F_{\text{GAN}}$ can not be directly used to modulate network $\cC$ until they are spatially aligned with the input. 
Taking the depicted image in Figure~\ref{fig:framework} as an example, the inversion image $x_{inv}^{rgb}$ contains all the semantic components that appeared in $x^l$ (\ie, the hen, soil and the weeds), but it is not spatially aligned with the input image. 
 
Before we introduce the align operation, it is worth mentioning that the prior color features $F_{\text{GAN}}$ are multi-scale features, and the features from a certain scale are used to modulate the layers of the corresponding scale in $\cC$. For simplicity, we only use a specific scale in $F_{\text{GAN}}$ to illustrate the align operation, denoted as $F_{\text{GAN}}^s \in \mathbb{R}^{C\times H' \times W'}$. 

In the alignment module, we first use two feature extractors with a shared backbone (denoted as $\cF_{L\to S}$ and $\cF_{RGB\to S}$, respectively) to project $x^l$ and $x_{inv}^{rgb}$ to a shared feature space $S$, obtaining the feature maps $\cF_{L\to S}(x^{l})$ and $\cF_{RGB\to S}(x_{inv}^{rgb})$.
After that, we use a non-local operation~\cite{Wang_2018} to calculate the correlation matrix $\cM \in \mathbb{R}^{H'W'\times H'W'}$ between the two feature maps. $\cM(u, v)$ denotes the similarity between the input and the GAN features in position $u$ and $v$. This operation is widely adopted to calculate dense semantic correspondence between two feature maps~\cite{lee2020reference,Zhang_2019,Zhang_2020}. 
Finally, we use the correlation matrix $\cM$ to warp $F_{\text{GAN}}^s$ and obtain the aligned GAN features at scale $s$, which are then used to modulate corresponding layers in $\cC$ at scale $s$.

\subsection{Objectives}\label{sec:losses}
\noindent\textbf{GAN inversion losses.}
As we have the ground truth colorful images $x^{rgb}$ during training, we can train GAN encoder $\cE$ by minimizing the difference between $x_{inv}^{rgb}$ and $x^{rgb}$. Instead of directly constraining the pixel values of $x_{inv}^{rgb}$ and $x^{rgb}$, we choose to minimize the discrepancy between their features extracted by the pre-trained discriminator $\cD^g$ of $\cG$, as this will make the GAN inversion easier to optimize (especially for BigGAN)~\cite{daras2020your,pan2020dgp}.
Formally, the discriminator feature loss $\cL_{inv\_ftr}$ is defined as:
\begin{equation}
\cL_{inv\_ftr} = \sum_l \norm{\cD_l^g(x_{inv}^{rgb})-\cD_l^g(x^{rgb})}_1,
\end{equation}
where $\cD_l^g$ represents the feature map extracted at $l$-th layer from $\cD^g$. 

To keep $z$ within a reasonable range, we also add a $L_2$ norm $\cL_{inv\_reg}$, otherwise, the GAN inversion with grayscale input will lead to unstable results:
\begin{equation}
\cL_{inv\_reg} = \frac{1}{2}\norm{z}_2.
\end{equation}

\noindent\textbf{Adversarial loss.}
We adopt the adversarial loss to help the colorization images looking more realistic. To stabilize the adversarial training, we employ the loss functions in LSGAN~\cite{Mao_2017}:
\begin{equation}
\begin{split}
\cL^{\cD}_{adv} &= \Ebb [(\cD^c(x^{lab})-1)^2] + \Ebb [(\cD^c(\hat{x}^{lab}))^2],\\
\cL^{\cG}_{adv} &= \Ebb [(\cD^c(\hat{x}^{lab})-1)^2],
\end{split}
\end{equation}
where $\cD^{C}$ is the discriminator to discriminate the colorization images $\hat{x}^{lab}$ generated from $\cC$ and color images $x^{lab}$ from real world. $\cC$ and $\cD^{C}$ are trained alternatively with $\cL^{\cG}_{adv}$ and $\cL^{\cD}_{adv}$, respectively.

\noindent\textbf{Perceptual loss.}
To make the colorization image perceptual plausible, we use the perceptual loss introduced in \cite{Johnson_2016}:
\begin{equation}
    \cL_{perc} = \norm{\phi_l(\hat{x}^{rgb})-\phi_l(x^{rgb})}_2,
\end{equation}
where $\phi_l$ represents the feature map extracted at $l$-th layer from a pretrained VGG19 network. Here ,we set $l=relu5\_2$.

\noindent\textbf{Domain alignment loss.}
To ensure that the grayscale image and inversion image are mapped to a shared feature space in correlation calculation, we adopt a domain alignment loss~\cite{Zhang_2020}:
\begin{equation}
\cL_{dom} = \norm{\cF_{L\to S}(x^{l})-\cF_{RGB\to S}(x^{rgb})}_1.
\end{equation}

\noindent\textbf{Contextual Loss.}
The contextual loss~\cite{Mechrez_2018} can measure the similarity between two non-aligned images. This loss term was originally designed for image translation tasks with non-aligned data, and then extended to exemplar-based colorization~\cite{Zhang_2019} to encourage colors in the output image to be close to those in the reference image. 
In this paper, we also adopt this loss to encourage the colorization image $\hat{x}^{rgb}$ to be relevant to the inversion image $x_{inv}^{rgb}$. Formally, the contextual loss $\cL_{ctx}$ is defined as:
\begin{equation}
    \cL_{ctx} = \sum_l \omega_{l}\left(-\log(\mathtt{CX}(\phi_l(\hat{x}^{rgb}), \phi_l(x_{inv}^{rgb}))) \right),
\end{equation}
where $\mathtt{CX}$ denotes the similarity metric between two features and we refer readers to \cite{Mechrez_2018} for more details. We use the layer $l=relu\{3\_2,4\_2,5\_2\}$ from the pretrained VGG19 network with weight $\omega_{l}=\{2,4,8\}$ to calculate $\cL_{ctx}$. 
We set larger loss weights for deeper features to guide the network to focus more on the semantic similarity instead of pixel-wise color closeness.

\noindent\textbf{Full objective.}
The full objective to train the GAN encoder $\cE$ is formulated as
\begin{equation}
\cL_{\cE} = \lambda_{inv\_ftr}\cL_{inv\_ftr} + \lambda_{inv\_reg}\cL_{inv\_reg}.
\end{equation}
The full objective to train the colorization network $\cC$ is formulated as
\begin{equation}
\begin{split}
\cL_{\cC} = &\lambda_{dom}\cL_{dom} + \lambda_{perc}\cL_{perc}\\ &+\lambda_{ctx}\cL_{ctx} + \lambda_{adv}\cL^{\cG}_{adv}.
\end{split}
\end{equation}
The full objective to train the discriminator $\cD_{C}$ for colorization is formulated as
\begin{equation}
\cL_{\cD_{C}} = \lambda_{adv}\cL^{\cD}_{adv}.
\end{equation}
In the above loss objectives, $\lambda_{inv\_ftr}$, $\lambda_{inv\_reg}$, $\lambda_{dom}$,$\lambda_{perc}$, $\lambda_{ctx}$ and $\lambda_{adv}$ are hyper-parameters that control the relative importance of each loss term.

\section{Experiments}\label{sec:experiments}

\subsection{Implementation Details}
\label{sec:exp:implementation}
We conduct our experiments on ImageNet~\cite{Russakovsky_2015}, a multi-class dataset, with a pretrained BigGAN~\cite{brock2018large} as the generative color prior.
We train our GCP-Colorization with the official training set. All the images are resized to $256\times 256$.

Our training process consists of two stages.
In the first stage, we freeze the pretrained BigGAN generator and train a BigGAN encoder with $\cL_{\cE}$. We utilize the last three layers of BigGAN discriminator to calculate the feature loss, where $\lambda_{inv\_ftr}$ is set to $1.0$ and $\lambda_{inv\_reg}$ is set to $0.0125$. We train the model for totally $10$ epochs with an initial learning rate of $1$e$-5$.

In the second stage, the BigGAN encoder is frozen and the rest of networks are trained end-to-end. We set the $\lambda_{perc}$ to $1$e$-3$, $\lambda_{adv}$ to $1.0$, $\lambda_{dom}$ to $10.0$ and $\lambda_{ctx}$ to $0.1$. We train the model for totally $10$ epochs with an initial learning rate of $1$e$-4$.
In both two stages, we set the batch size to $8$ per GPU and decay the learning rate linearly to $0$. All the models are trained with the Adam~\cite{Kingma_2015} optimizer.

The BigGAN encoder has a similar structure to the BigGAN discriminator, while the class embedding is injected into the BigGAN encoder following the BigGAN generator.
To reduce the learnable parameters of $\cF_{L\to S}$ and $\cF_{RGB\to S}$, their backbones are shared.
The discriminator for colorization $\cD^c$ is a three-scale PatchGAN network~\cite{Isola_2017,Wang_2018} with spectral normalization~\cite{Miyato_2018}. 

\subsection{Comparisons with Previous Methods}
\label{sec:exp:comparison}
\begin{figure*}[ht]
    \centering
    \vspace{-0.4cm}
    \includegraphics[width=\textwidth]{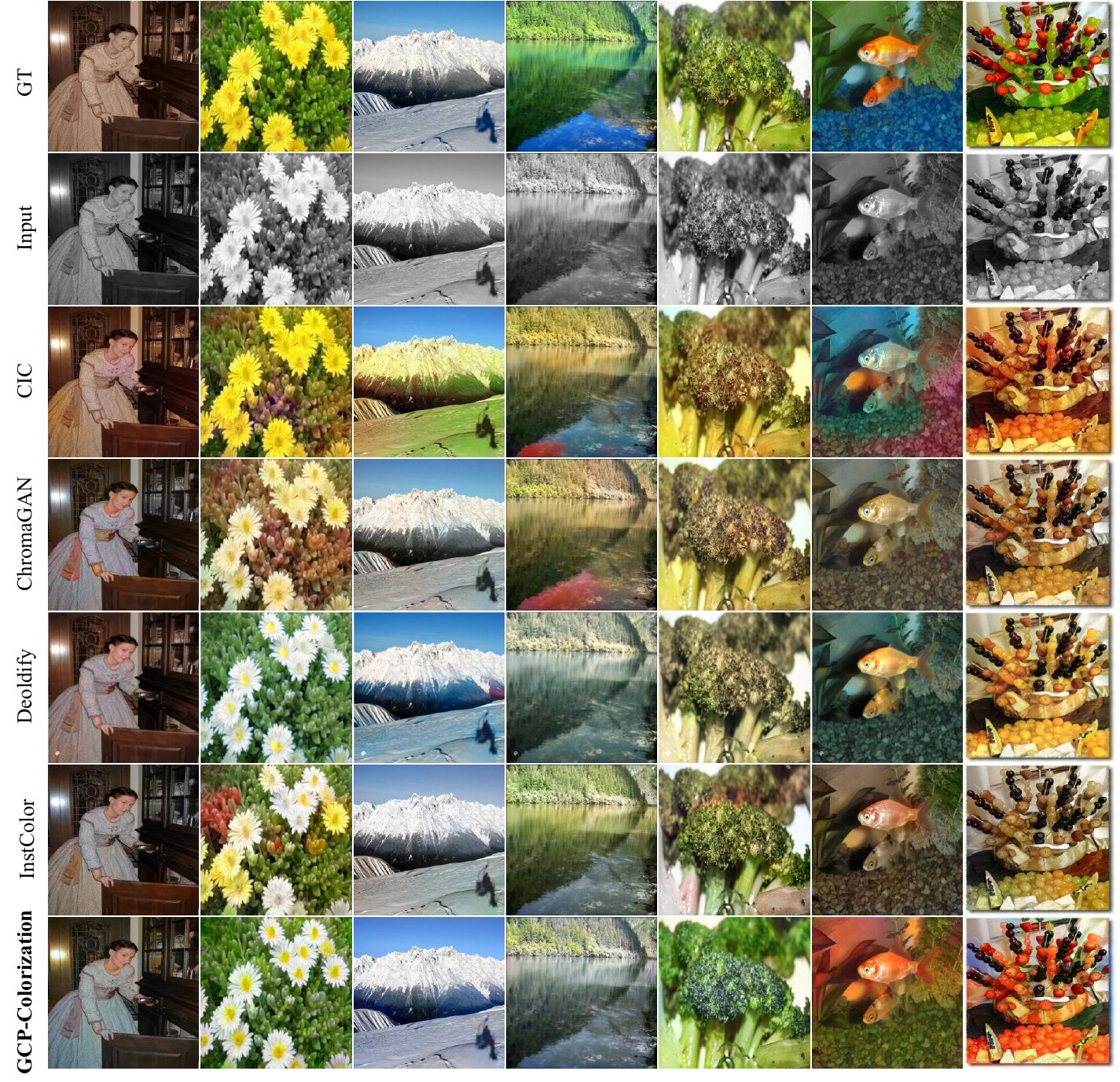}
    \vspace{-0.6cm}
    \caption{Visual comparisons with previous automatic colorization methods. Our GCP-Colorization is able to generate more natural and vivid colors.}
    \label{fig:v_com}
    \vspace{-0.4cm}
\end{figure*}

To evaluate the performance of our method, we compare GCP-Colorization to other state-of-the-art automatic colorization methods including CIC~\cite{Zhang_2016}, ChromaGAN~\cite{Vitoria_2020}, DeOldify~\cite{DeOldify} and InstColor~\cite{Su_2020}. We re-implement CIC and ChromaGAN in PyTorch based on the original paper.

\noindent\textbf{Quantitative comparison.} 
We test all the methods on the ImageNet validation set of 50,000 images.
The quantitative results on five metrics are reported in Table~\ref{table:quantitative}. 
\textbf{Fr\'echet Inception Score (FID)}~\cite{heusel2017gans} measures the distribution similarity between the colorization results and the ground truth color images. Our GCP-Colorization achieves the lowest FID, indicating that GCP-Colorization could generate colorization results with better image quality and fidelity. \textbf{Colorfulness Score}~\cite{hasler2003measuring} reflects the vividness of generated images. 
Our GCP-Colorization obtains a better colorfulness score (2nd column) and closest colorfulness score to the ground-truth images (3rd column).
The highest colorfulness score of CIC is an outlier, also observed in~\cite{Zhang_2019}. This is probably because CIC encourages rare colors in the loss function, and the rare colors are misjudged as vivid colors by this metric.
We also provide PSNR and SSIM for reference. However, it is well-known that such pixel-wise measurements may not well reflect the actual performance~\cite{cao2017unsupervised, He_2018,Messaoud_2018,Su_2020,Vitoria_2020,Zhao_2019}, as plausible colorization results probably diverge a lot from the original color image.

\begin{table}[!th]
	\footnotesize
    \centering 
    \setlength\tabcolsep{5  pt}
    \caption{Quantitative comparison. $\Delta$Colorful denotes the absolute colorfulness score difference between the colorization images and the ground truth color images.} 
    \begin{tabular}{@{}lccccc@{}}
    \toprule
    & FID$\downarrow$ & Colorful$\uparrow$ & $\Delta$Colorful$\downarrow$ & PSNR$\uparrow$ & SSIM$\uparrow$\\
    \midrule
      CIC   & 19.71 & \textbf{43.92} & 5.57 & 20.86 &0.86 \\
		ChromaGAN &5.16 &27.49 & 10.86 & 23.12 & 0.87\\
		DeOldify &3.87 &22.83 & 15.52 & 22.97 & 0.91\\
		InstColor &7.36 &27.05 & 11.30 & 22.91 & 0.91\\
		\textbf{GCP-Colorization} &\textbf{3.62} &35.13 &\textbf{3.22} &21.81 &0.88 \\
      \bottomrule
    \end{tabular}
    \vspace{-0.4cm}
    \label{table:quantitative}
\end{table}

\noindent\textbf{Qualitative Comparison.}
As shown Figure~\ref{fig:v_com}, compared to CIC, ChromaGAN and Deoldify, our GCP-Colorization tend to obtain more natural and vivid colors.
For instance, in contrast to our vivid green, the broccoli color (Column 5) of the other methods looks unnatural and yellowish.
In addition, GCP-Colorization generates better results with regards to consistent hues.
In column 1, we can observe inconsistent chromaticities in the results of Deoldify and InstColor, where the dress is yellow and blue at the same time. Though ChromaGAN has consistent tones, it fails to capture the details by only mapping overall blue tones to the image, leading to the blue artifacts on the lady's face.
Instead, GCP-Colorization successfully maintains the consistent tone and captures the details as shown in Column 1.
Furthermore, our GCP-Colorization can yield more diverse and lively colors for each small instance as shown in column 7, while InstColor fails to detect such small objects and inherently degrades to use its global image colorization network, whose results are dim and less vivid.

\noindent\textbf{User study.}
\begin{figure}
    \centering
    \vspace{-0.4cm}
    \includegraphics[width=0.95\linewidth]{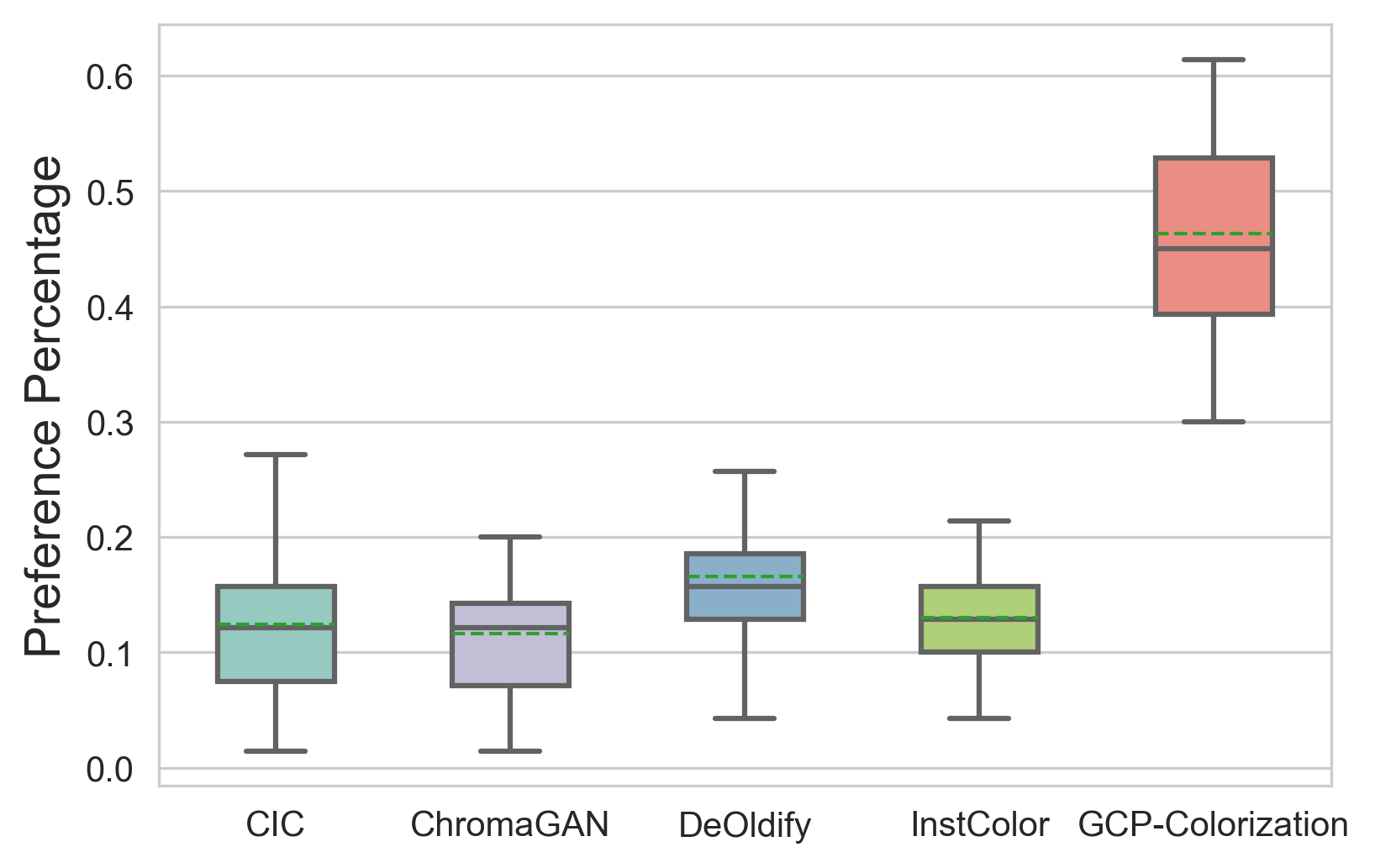}
    \vspace{-0.4cm}
    \caption{Boxplots of user preferences for different methods. Green dash lines represent the means. Our GCP-Colorization got a significantly higher preference rate by users than other colorization methods.}
    \label{fig:user_study}
    \vspace{-0.4cm}
\end{figure}
In order to better evaluate the subjective quality (\ie., vividness and diverseness of colors), we conduct a user study to compare our GCP-Colorization with the other state-of-art automatic colorization methods, \ie., CIC~\cite{Zhang_2016}, ChromaGAN~\cite{Vitoria_2020}, DeOldify~\cite{DeOldify} and InstColor~\cite{Su_2020}. 
We recruited 30 participants with normal or corrected-to-normal vision and without color blindness.
We randomly select 70 images from various categories of image contents (\eg, animals, plants, human beings, landscape, food, \etc).
For each image, the grayscaled image is displayed at the leftmost while five colorization results are displayed in a random manner to avoid potential bias.
The participants are required to select the best-colorized image according to the colorization quality in terms of vividness and diverseness.
The boxplots are displayed in Figure~\ref{fig:user_study}.
Our GCP-Colorization obtained a significant preference ($~46.3\%$) by users than other colorization methods (CIC $12.4\%$, ChromaGAN $11.7\%$, Deoldify $16.6\%$, InstColor $13.0\%$), demonstrating distinct advantage on producing natural and vivid results.

\subsection{Ablation Studies}
\noindent\textbf{Generative color prior.}
\begin{figure}
    \centering
    \vspace{-0.6cm}
    \includegraphics[width=\linewidth]{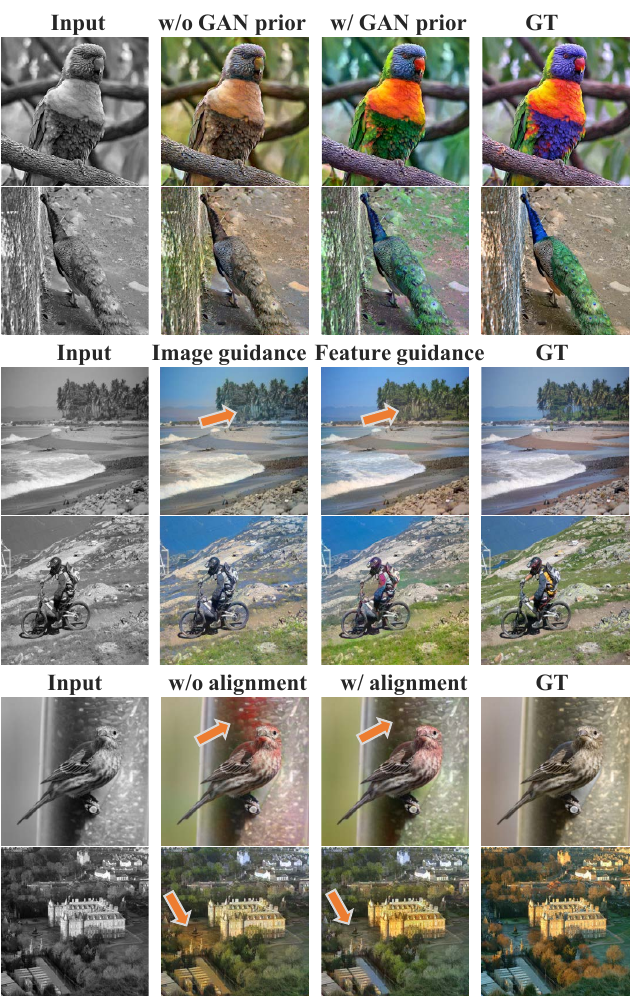}
    \vspace{-0.6cm}
    \caption{Qualitative comparisons of ablation studies on GAN prior, feature guidance and spatial alignment. GAN prior introduces brighter and more various colors. Multi-resolution feature guidance could produce more coherent and rich colors than image guidance. Spatial alignment mitigates the misalignment between `retrieved' GAN features and input gray images.}
    \vspace{-0.4cm}
    \label{fig:ablation}
\end{figure}
Generative color prior plays an important role in providing exemplar features for vivid and various colors. 
When we remove the pretrained GANs, our method degrades to a common automatic colorization method without guidance. 
As shown in Figure~\ref{fig:ablation}, the variant experiment without GAN prior generates dim and bleak colors, while our method could produce bright and joyful images, under the guidance of colorful exemplars in GAN distributions.
We could also observe a large drop in FID and Colorfulness score without generative color prior (Table~\ref{table:ablation}).

\noindent\textbf{Feature guidance \textit{vs.} image guidance.}
Our method adopts multi-resolution intermediate GAN features to guide the colorization.
The design could fully facilitate the color and detail information from GAN features. 
Compared to image guidance, our results are capable of recovering more vivid colors faithful to their underlying classes (\eg, the trees on the beach, the grass on the mountains), as multi-resolution features contain more semantic and color information (Figure~\ref{fig:ablation} and Table~\ref{table:ablation}). 

\noindent\textbf{Spatial alignment.}
The prior features from pretrained GANs probably misalign with the input images. To address this problem, we employ spatial alignment to align the GAN features, leading to fewer artifacts and coherent results for colorization. As shown in Figure~\ref{fig:ablation}, the tree and square colors of our results are not affected by other objects. 

\begin{table}[!t]
	\footnotesize
    \centering 
    \setlength{\tabcolsep}{10 pt}
    \vspace{-0.4cm}
    \caption{Quantitative comparisons for ablation studies. $\Delta$Colorful denotes the absolute colorfulness score difference between the colorization images and the ground truth color images.}
    \vspace{-0.2cm}
    \begin{tabular}{@{}lccc@{}}
    \toprule
    Variants & FID$\downarrow$ & Colorful$\uparrow$ & $\Delta$Colorful$\downarrow$\\
    \midrule
    Full Model &\textbf{3.62} &\textbf{35.13} &\textbf{3.22}  \\
    w/o Generative Color Prior   & 8.40 & 31.21 & 7.14  \\
    Image Guidance &4.01 &26.12 & 12.23 \\
    w/o Spatial Alignment & 4.59 & 31.94 & 6.41 \\
    \bottomrule
    \end{tabular}
    \label{table:ablation}
\end{table}

\subsection{Controllable Diverse Colorization}
\begin{figure}
    \centering
    \vspace{-0.4cm}
    \includegraphics[width=\linewidth]{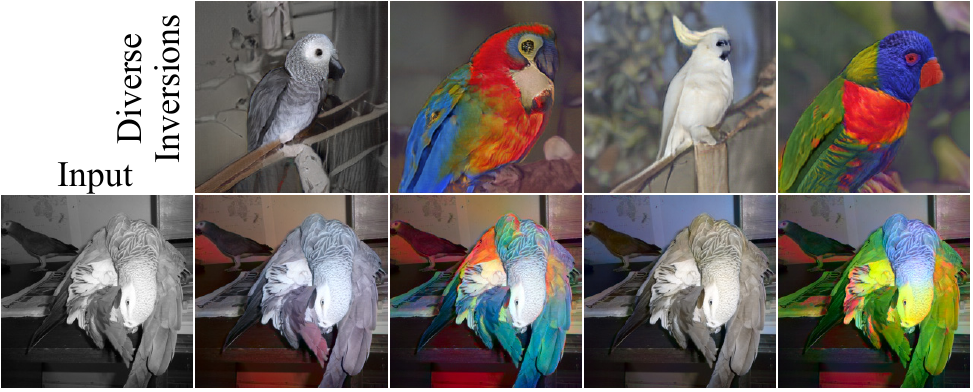}
    \vspace{-0.6cm}
    \caption{Our method could adjust the latent codes to obtain various inversion results, thus easily achieving diverse colorization results for the parrot.}
    \label{fig:ab_diverse}
    \vspace{-0.4cm}
\end{figure}

In order to achieve diverse colorization, many attempts have been made. For automatic colorization methods, diversity is usually achieved by stochastic sampling~\cite{Deshpande_2017,Guadarrama_2017,Royer_2017}. However, the results are barely satisfactory and difficult to control (Detailed comparisons are provided in the supplementary material). For exemplar-based colorization methods~\cite{He_2018,Xu_2020}, it is time-consuming and challenging to find large amounts of reference images with various styles. Different from previous methods, our GCP-Colorization could achieve diverse and controllable colorization in a much easier and novel manner.

\begin{figure}
    \centering
    \vspace{-0.6cm}
    \includegraphics[width=\linewidth]{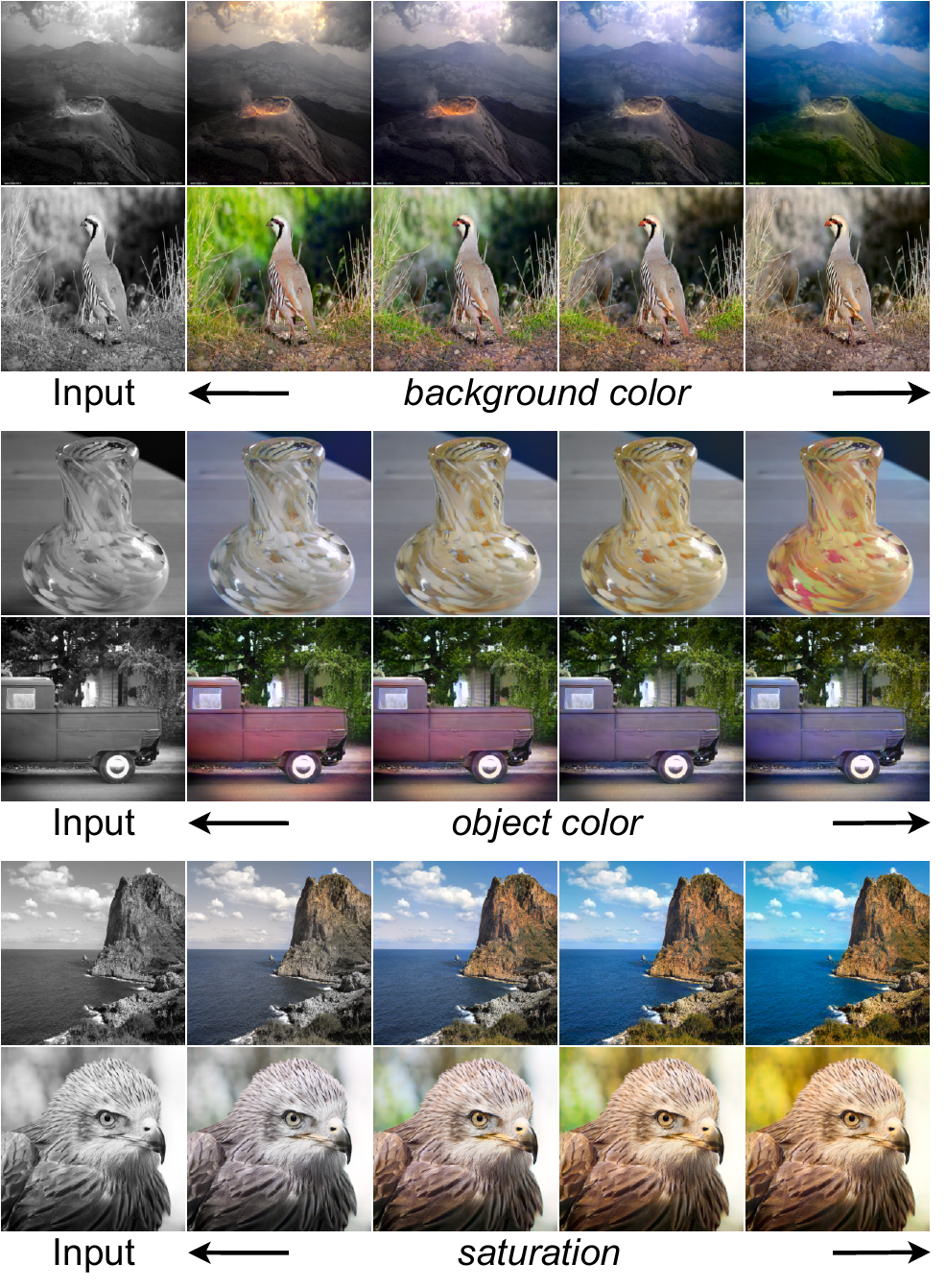}
    \vspace{-0.7cm}
    \caption{With the interpretable controls of GANs, our method could attain controllable and smooth transitions by walking through GAN latent space.}
    \vspace{-0.3cm}
    \label{fig:ab_div_2}
\end{figure}

On the one hand, we can adjust the latent codes with disturbance to obtain various exemplars and their corresponding features. 
As shown in Figure~\ref{fig:ab_diverse}, various inversion results are attained by adjusting the latent codes, and correspondingly, diverse colorization results of parrots are obtained. 
On the other hand, GCP-Colorization inherits the merits of interpretable controls of GANs~\cite{erik2020ganspace,gansteerability,shen2021closedform,Voynov_2020} and could attain controllable and smooth transitions by walking through GAN latent space. 
Specifically, we employ an unsupervised method~\cite{Voynov_2020} to find color-relevant directions, such as lighting, saturation, \etc. As depicted in Figure~\ref{fig:ab_div_2}, smooth transitions on background colors, object colors, saturation could be achieved. 

\subsection{Limitations}
\begin{figure}
    \centering
    \vspace{-0.2cm}
    \includegraphics[width=\linewidth]{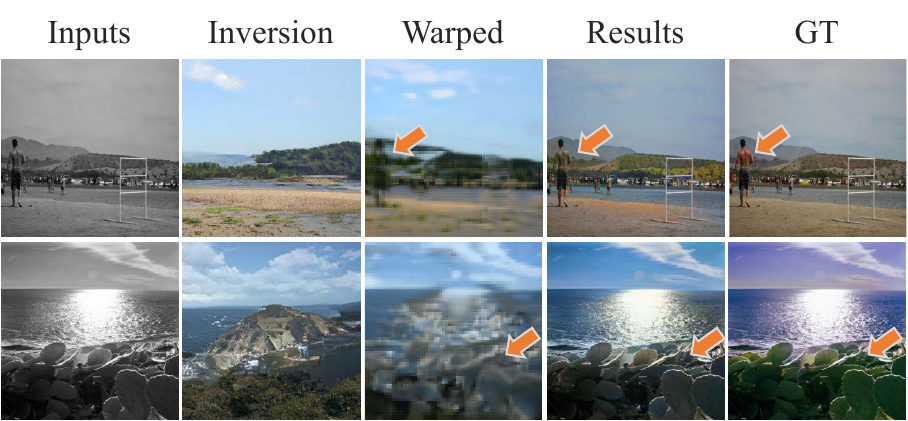}
    \vspace{-0.6cm}
    \caption{Limitations of our model. The human in the beach and the cactus are missing from the GAN inversion, resulting in unnatural colors.}
    \label{fig:limitation}
    \vspace{-0.6cm}
\end{figure}

Though our GCP-Colorization could produce appealing results in most cases, it still has limitations.
When the input image is not in the GAN distribution or GAN inversion fails, our method degrades to common automatic colorization methods and may result in unnatural and incoherent colors. 
As shown in Figure~\ref{fig:limitation}, the human on the beach and the cactus are missing from the GAN inversion results. Thus, our method cannot find corresponding color guidance, resulting in unnatural colorization for these objects.
This could be mitigated by improving the GAN inversion and common automatic colorization approaches in future works.

\section{Conclusion}
\vspace{-0.2cm}
In this work, we have developed the GCP-Colorization framework to produce vivid and diverse colorization results by leveraging generative color priors encapsulated in pretrained GANs.
Specifically, we `retrieve' multi-resolution GAN features conditioned on the input grayscale image with a GAN encoder, and incorporate such features into the colorization process with feature modulations.
One could easily realize diverse colorization by simply adjusting the latent codes or conditions in GANs.
Moreover, our method could also attain controllable and smooth transitions by walking through the GAN latent space. 

{\small
\bibliographystyle{ieee_fullname}
\bibliography{arXiv}
}

\newpage

\begin{center}
\textbf{{\Large Supplementary}}
\end{center}
\setcounter{section}{0}

\section{GAN Inversion Results}
\label{sec:inv_imgs}
 \begin{figure*}
     \centering
     \includegraphics[width=\linewidth]{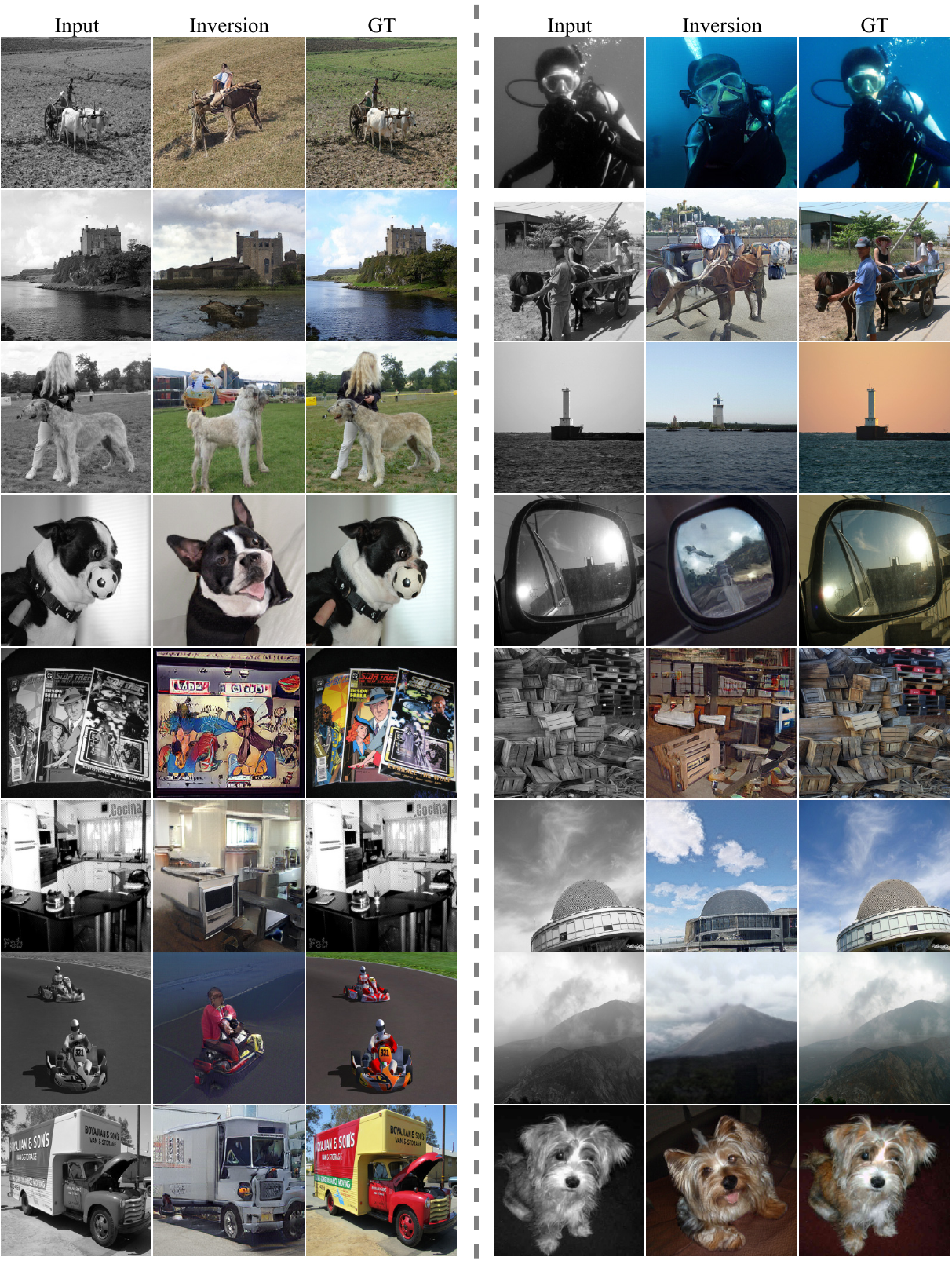}
     \caption{Visual results of BigGAN inversion. It is observed that the GAN inversion could generate colorful images that share the similar semantic contents to the gray inputs. \textbf{Left Column}: input gray-scale images. \textbf{Middle Column}: the GAN inversion results. \textbf{Right Column}: ground-truth colorful images.}
     \label{fig:more_inv}
 \end{figure*}
 
 We show the GAN inversion results in Fig.~\ref{fig:more_inv}. It is observed that the GAN inversion could generate colorful images that share the similar semantic contents to the gray inputs.

 \section{Ablation Study of BigGAN Inversion}
 \label{sec:ab_inv}
 Despite great success of GAN inversion in StyleGAN\cite{Zhu_2020}, optimizing an encoder-based BigGAN inversion model is still a challenging  task~\cite{daras2020your,pan2020dgp}. Simply optimizing the latent code $z$ leads to large $L_2$ norm of latent code $z$ and low-quality inverted images. We conjecture that this is because the self-attention architecture in BigGAN brings optimization challenges. 
  To tackle this issue, we directly add an extra  $L_2$ norm penalty to the latent code and empirically found that it could alleviate this issue and generate plausible inversion results.
 Besides, We also employ the discriminator feature loss $\cL_{inv\_ftr}$ instead of the perceptual loss trained on classification task, since the feature space of pretrained discriminator is more coherent with that of inversion images.
 Recently important progress has been achieved in StyleGAN inversion task, some useful tricks are introduced to generate high-quality inversion images. However, not all of those useful tricks are suitable for BigGAN, \eg, we found in practice that the domain-guided encoder introduced in \cite{Zhu_2020} will not bring benefits in BigGAN. Hence, the inversion of the encoder-based BigGAN still remains a challenge.
 
\begin{figure*}[h]
      \centering
      \includegraphics[width=0.95\linewidth]{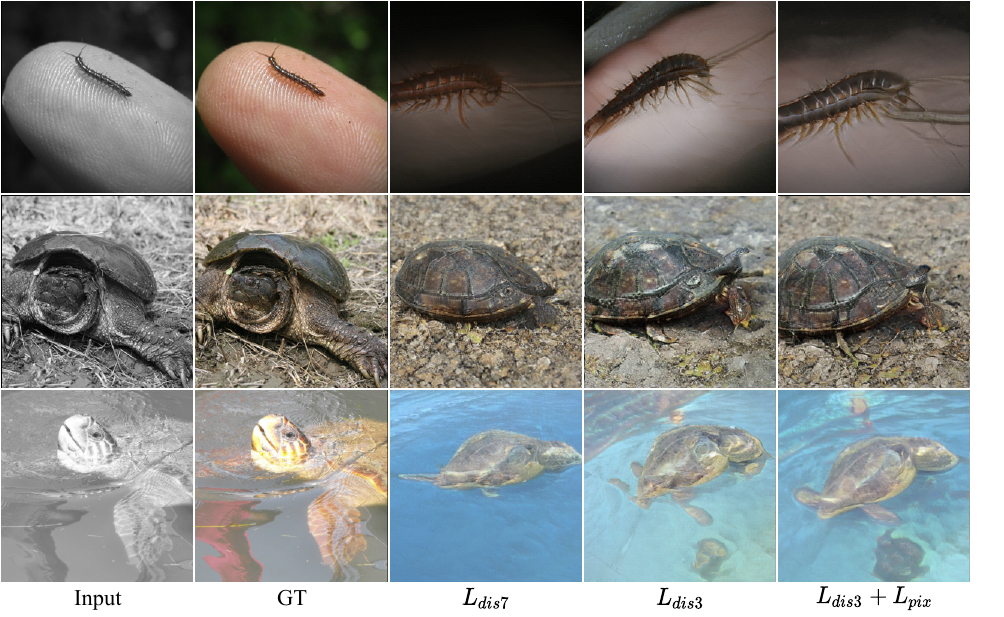}
      \caption{Ablation study of loss functions in BigGAN inversion. $\cL_{dis7}$ denotes using more discriminator layers (\ie, the last $7$ layers) to calculate the feature loss. $\cL_{dis3}$ denotes using the last $3$ discriminator layers to calculate the feature loss. $\cL_{dis3}+\cL_{pix}$ denotes using the last $3$ discriminator layers and pixel loss. It is observed that neither adding more discriminator layers nor introducing the pixel loss will generate better inversion images, but result in blurry images with fewer details. Therefore, we choose $\cL_{dis3}$ as our default setting.}
      \label{fig:ab_inversion}
  \end{figure*}

  In addition to the discriminator feature loss and $L_2$ norm scheme, we also make some attempts to improve BigGAN inversion but their results are no better than the aforementioned method. Those attempts include adopting more discriminator layers to calculate the feature loss and introducing pixel loss. 
  The comparisons of these attempts can be found in Fig.~\ref{fig:ab_inversion}. It is observed that neither adding more discriminator layers nor introducing the pixel loss will generate better inversion images, but result in blurry images with fewer details.

\section{Comparison with GAN Inversion}
\label{sec:com_inv}
\begin{figure*}[t]
    \centering
    \vspace{-4cm}
    \includegraphics[width=\linewidth]{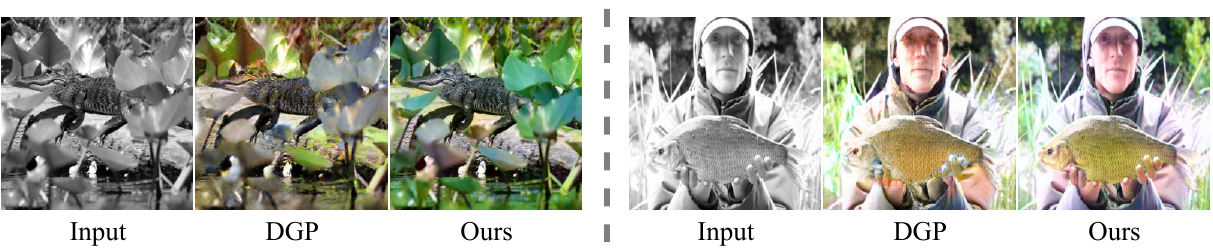}
    \caption{Comparisons with DGP.}
    \label{fig:com_inv}
\end{figure*}

As GAN inversions also incorporate pretrained GANs as a prior to guide the colorization process, we further compare the results of our GCP-Colorization against results of a GAN inversion method - DGP~\cite{pan2020dgp}.
Selected representative cases are presented in Figure~\ref{fig:com_inv} for a qualitative comparison.
As shown in the figure, we can observe that color boundaries are not clearly separated in the results of DGP.
The blurred color boundaries produced by DGP are inevitable considering the information loss during GAN inversion process.
The low-dimensional latent codes generated in DGP fail to encode the spatial information in the image, thereby DGP is neither capable of retaining the local details nor preserving the shape of objects.
On the other hand, our GCP-Colorization, which employs a SPADE modulation to incorporate the spatial features, has obtained better results in terms of texture faithfulness.
Apart from improving colorization quality, GCP-Colorization also reduces the computational cost by adopting straightforward feed-forward inference to avoid iterative optimization training process used in DGP.

\section{More Diverse Colorization Results}
\label{sec:diverse}
\begin{figure*}
    \centering
    \includegraphics[width=\linewidth]{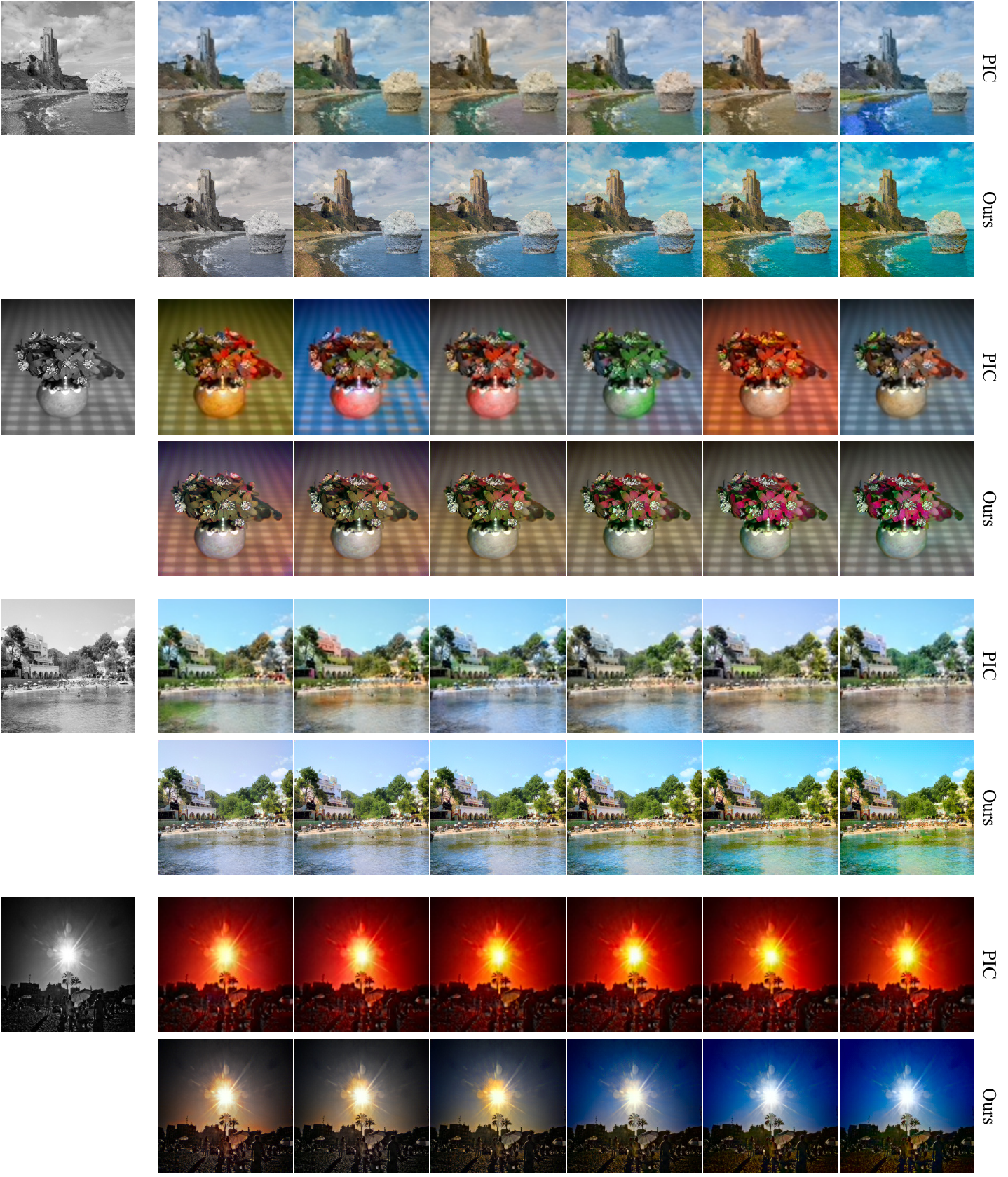}
    \caption{Comparisons of diverse colorization with PIC (\textbf{First row}: PIC; \textbf{Second row}: Ours). Zoom in to see the spatially incoherent and unrealistic colorizations from PIC. Our method could generate spatially coherent and more natural colorization results with smooth control.}
    \label{fig:div_color_1}
\end{figure*}

\begin{figure*}
    \centering
    \includegraphics[width=\linewidth]{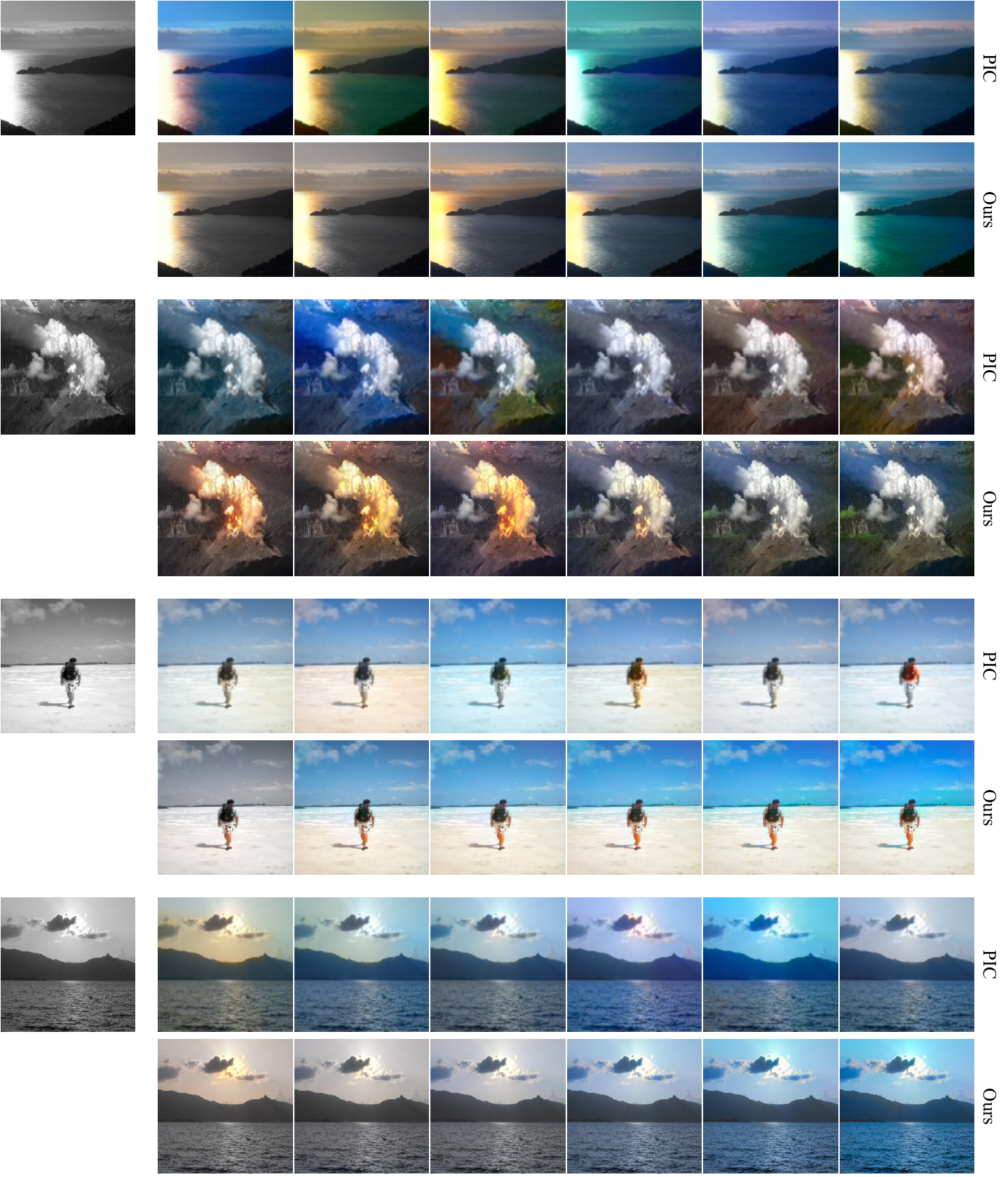}
    \caption{Comparisons of diverse colorization with PIC (\textbf{First row}: PIC; \textbf{Second row}: Ours). Zoom in to see the spatially incoherent and unrealistic colorizations from PIC. Our method could generate spatially coherent and more natural colorization results with smooth control.}
    \label{fig:div_color_2}
\end{figure*}

In this section, we provide comparisons of diverse colorization with stochastic sampling based method PIC~\cite{Royer_2017}. PIC adopts an autoregressive network PixelCNN to model the full joint distribution of pixel color values, by sampling from this distribution, diverse colorizations can be obtained. As shown in Fig.~\ref{fig:div_color_1} and Fig.~\ref{fig:div_color_2}, PIC usually generates \textbf{spatially incoherent} and \textbf{unrealistic} colorization results. On the contrary, our GCP-Colorization does not suffer from these issues. Besides, GCP-Colorization could attain smooth and controllable colorizations.

\section{User study results}
\label{sec:more_results}
\begin{figure*}[b]
	\centering
	\captionsetup{justification=centering}
	\includegraphics[width=\linewidth]{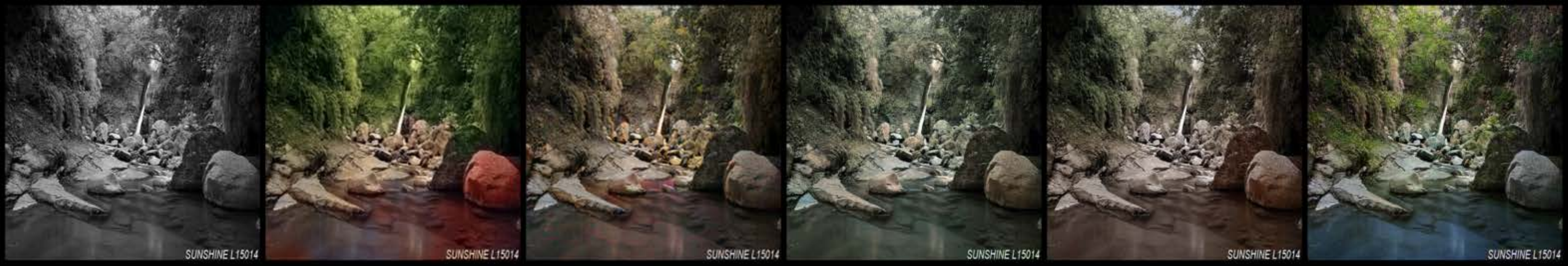}
	\caption{User preferences for different methods. \\From column 2-6: $10\%$(CIC) : $3\%$(ChromaGAN) : $13\%$(DeOldify) : $3\%$(InstColor) : $70\%$(GCP-Colorization).}
	\label{fig:us_first}
\end{figure*}

\begin{figure*}
	\centering
	\captionsetup{justification=centering}
	\includegraphics[width=\linewidth]{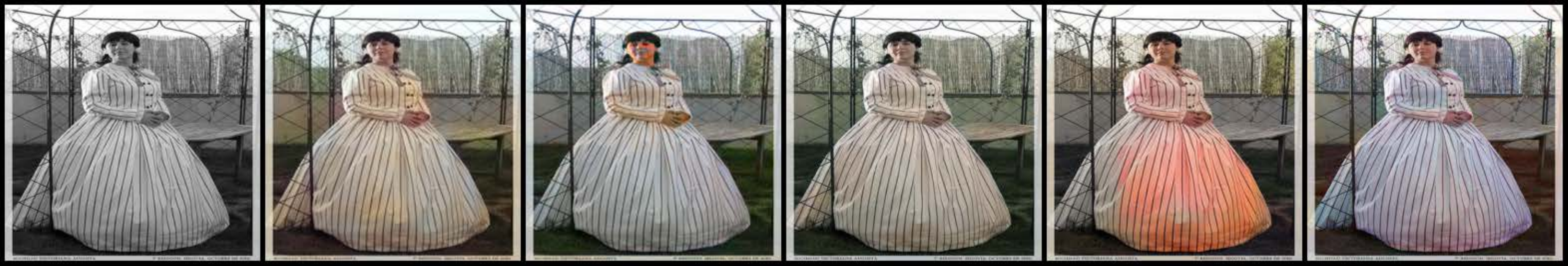}
	\caption{User preferences for different methods. \\From column 2-6: $17\%$(CIC) : $13\%$(ChromaGAN) : $13\%$(DeOldify) : $3\%$(InstColor) : $53\%$(GCP-Colorization).}
	\label{fig:us_last}
\end{figure*}

We show some cases from our user study in Fig.~\ref{fig:us_first} - \ref{fig:us_last}. 
Our GCP-Colorization got a significant preference  by users than other colorization methods (CIC~\cite{Zhang_2016}, ChromaGAN~\cite{Vitoria_2020}, DeOldify~\cite{DeOldify} and InstColor~\cite{Su_2020}), showing distinct advantage on producing natural and vivid results.

\begin{figure*}[b]
	\centering
	\captionsetup{justification=centering}
	\includegraphics[width=\linewidth]{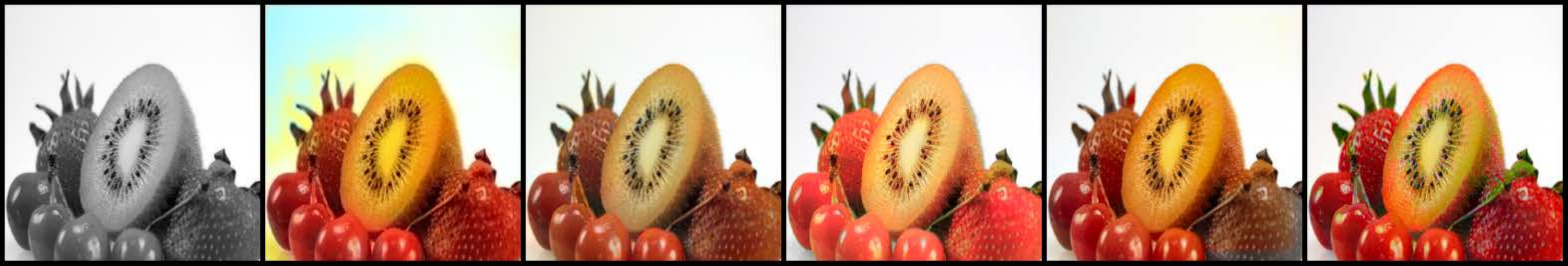}
	\caption{User preferences for different methods. \\From column 2-6: $10\%$(CIC) : $7\%$(ChromaGAN) : $23\%$(DeOldify) : $10\%$(InstColor) : $50\%$(GCP-Colorization).}
\end{figure*}

\begin{figure*}[b]
	\centering
	\captionsetup{justification=centering}
	\includegraphics[width=\linewidth]{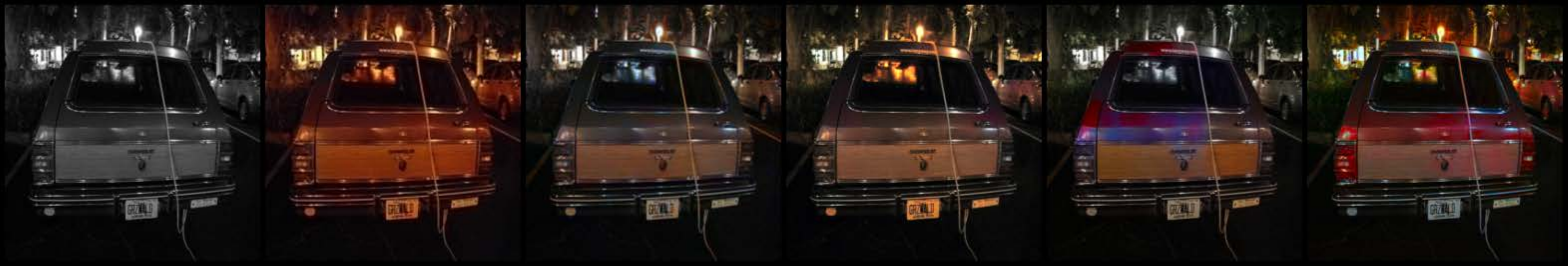}
	\caption{User preferences for different methods. \\From column 2-6: $3\%$(CIC) : $3\%$(ChromaGAN) : $30\%$(DeOldify) : $3\%$(InstColor) : $60\%$(GCP-Colorization).}
\end{figure*}

\begin{figure*}[b]
	\centering
	\captionsetup{justification=centering}
	\includegraphics[width=\linewidth]{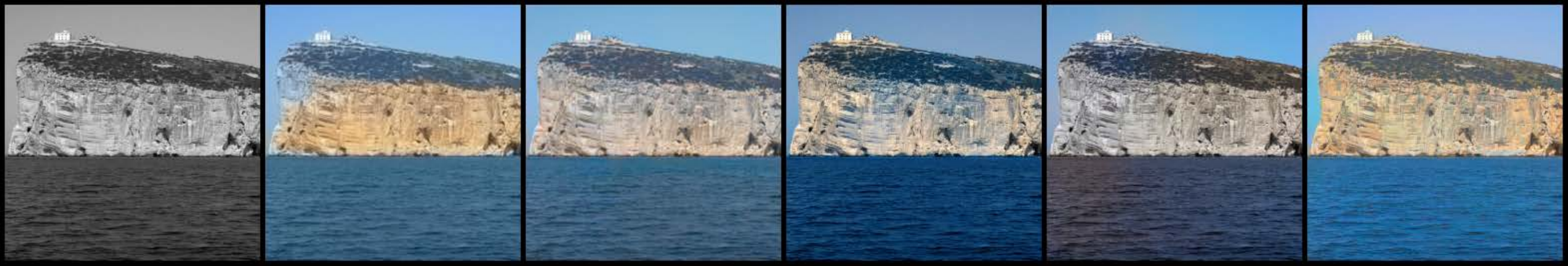}
	\caption{User preferences for different methods. \\From column 2-6: $7\%$(CIC) : $10\%$(ChromaGAN) : $20\%$(DeOldify) : $10\%$(InstColor) : $53\%$(GCP-Colorization).}
\end{figure*}

\begin{figure*}
	\centering
	\captionsetup{justification=centering}
	\includegraphics[width=\linewidth]{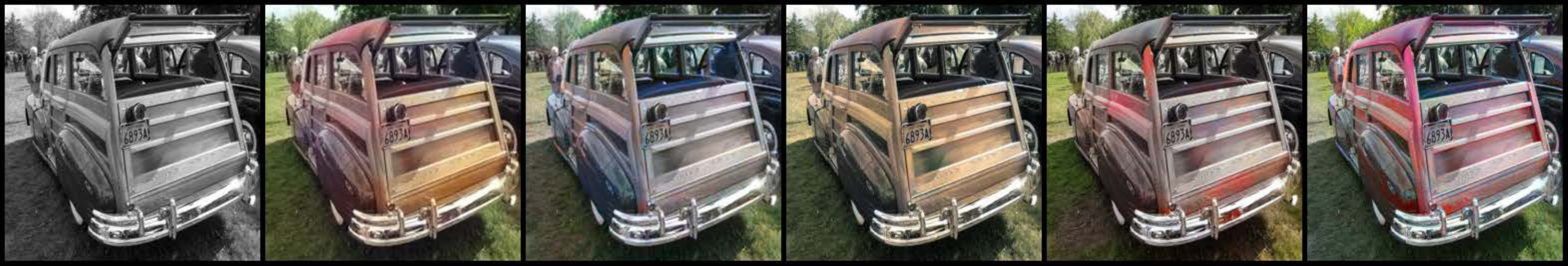}
	\caption{User preferences for different methods. \\From column 2-6: $0\%$(CIC) : $7\%$(ChromaGAN) : $0\%$(DeOldify) : $23\%$(InstColor) : $70\%$(GCP-Colorization).}
\end{figure*}

\begin{figure*}
	\centering
	\captionsetup{justification=centering}
	\includegraphics[width=\linewidth]{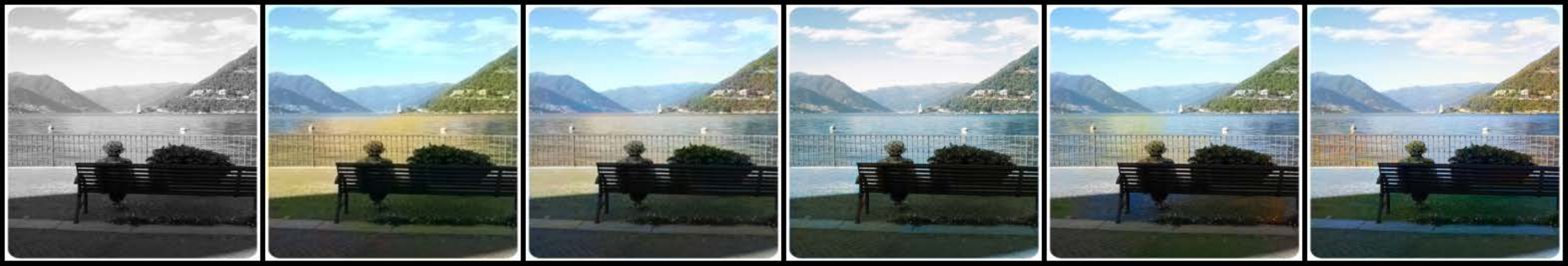}
	\caption{User preferences for different methods. \\From column 2-6: $7\%$(CIC) : $13\%$(ChromaGAN) : $17\%$(DeOldify) : $23\%$(InstColor) : $40\%$(GCP-Colorization).}
\end{figure*}

\begin{figure*}
	\centering
	\captionsetup{justification=centering}
	\includegraphics[width=\linewidth]{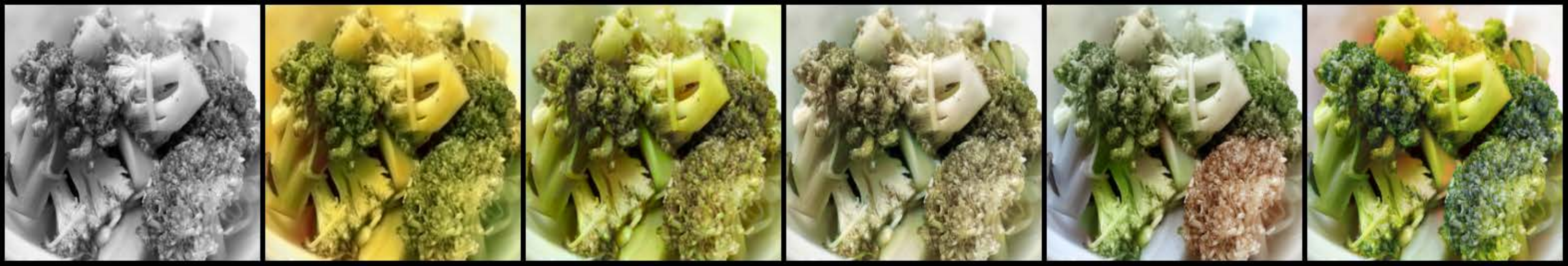}
	\caption{User preferences for different methods. \\From column 2-6: $0\%$(CIC) : $13\%$(ChromaGAN) : $3\%$(DeOldify) : $33\%$(InstColor) : $50\%$(GCP-Colorization).}
\end{figure*}

\begin{figure*}
	\centering
	\captionsetup{justification=centering}
	\includegraphics[width=\linewidth]{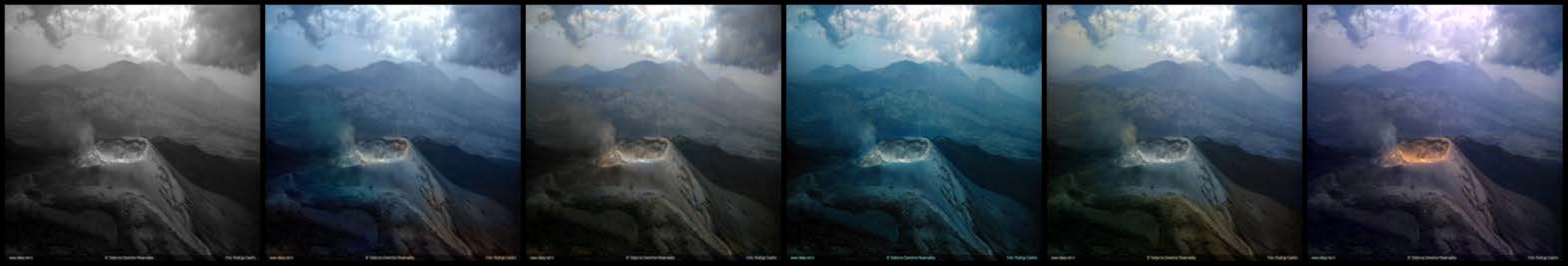}
	\caption{User preferences for different methods. \\From column 2-6: $17\%$(CIC) : $10\%$(ChromaGAN) : $27\%$(DeOldify) : $0\%$(InstColor) : $47\%$(GCP-Colorization).}
\end{figure*}

\begin{figure*}
	\centering
	\captionsetup{justification=centering}
	\includegraphics[width=\linewidth]{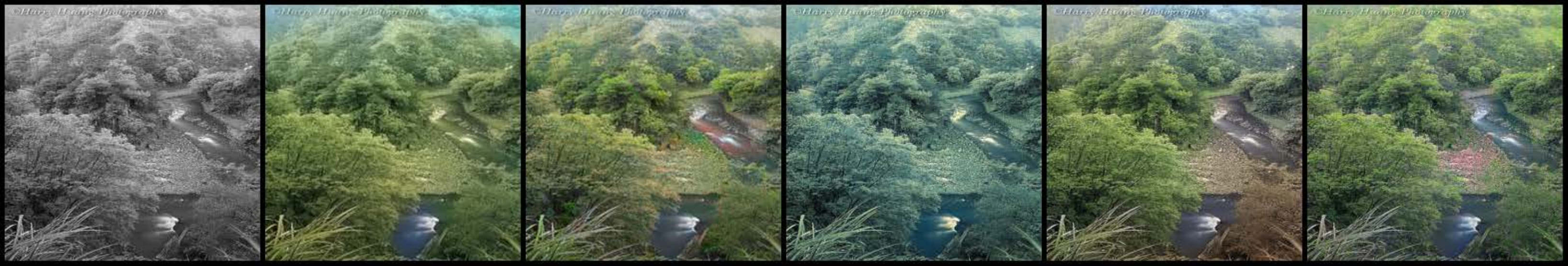}
	\caption{User preferences for different methods. \\From column 2-6: $7\%$(CIC) : $7\%$(ChromaGAN) : $0\%$(DeOldify) : $27\%$(InstColor) : $60\%$(GCP-Colorization).}
\end{figure*}

\begin{figure*}
	\centering
	\captionsetup{justification=centering}
	\includegraphics[width=\linewidth]{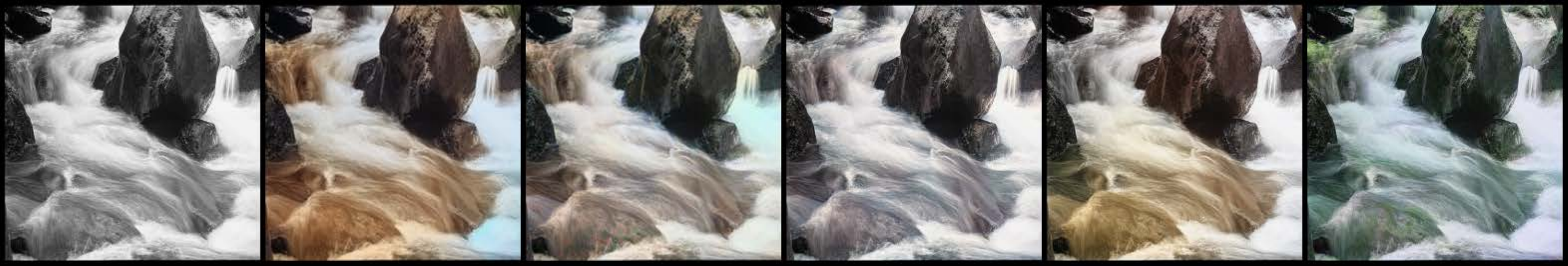}
	\caption{User preferences for different methods. \\From column 2-6: $3\%$(CIC) : $7\%$(ChromaGAN) : $13\%$(DeOldify) : $3\%$(InstColor) : $73\%$(GCP-Colorization).}
\end{figure*}

\begin{figure*}
	\centering
	\captionsetup{justification=centering}
	\includegraphics[width=\linewidth]{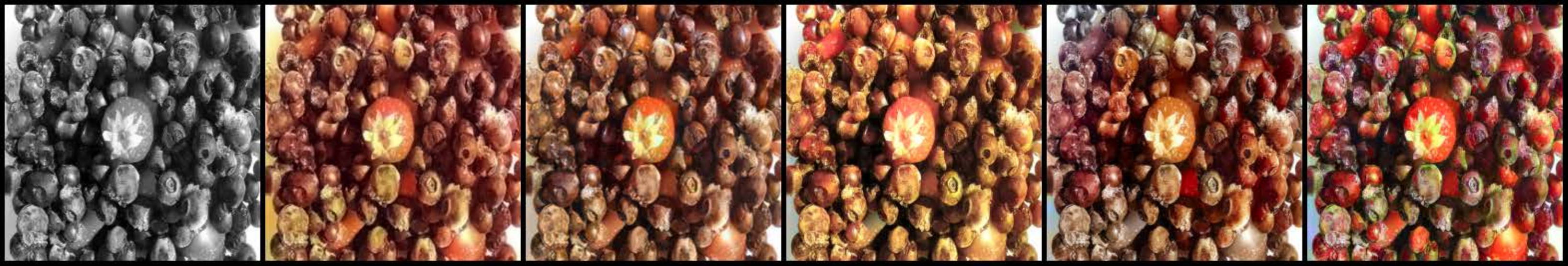}
	\caption{User preferences for different methods. \\From column 2-6: $13\%$(CIC) : $7\%$(ChromaGAN) : $0\%$(DeOldify) : $0\%$(InstColor) : $80\%$(GCP-Colorization).}
\end{figure*}

\begin{figure*}
	\centering
	\captionsetup{justification=centering}
	\includegraphics[width=\linewidth]{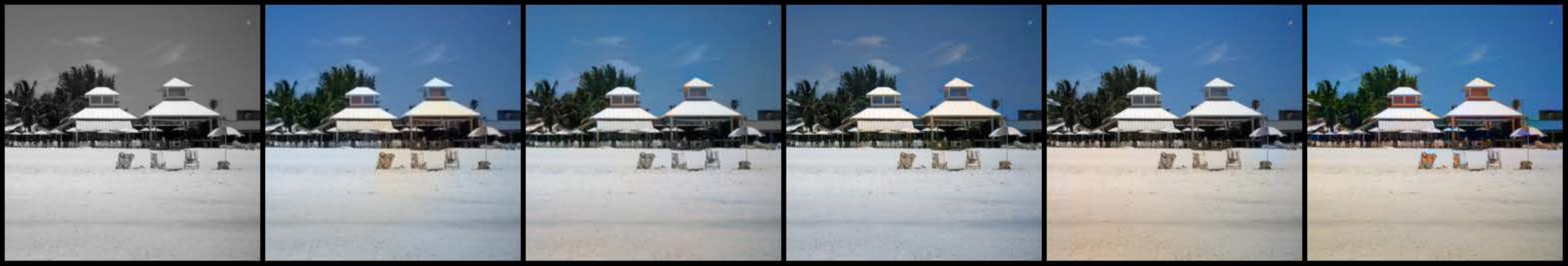}
	\caption{User preferences for different methods. \\From column 2-6: $7\%$(CIC) : $13\%$(ChromaGAN) : $17\%$(DeOldify) : $10\%$(InstColor) : $53\%$(GCP-Colorization).}
\end{figure*}

\begin{figure*}
	\centering
	\captionsetup{justification=centering}
	\includegraphics[width=\linewidth]{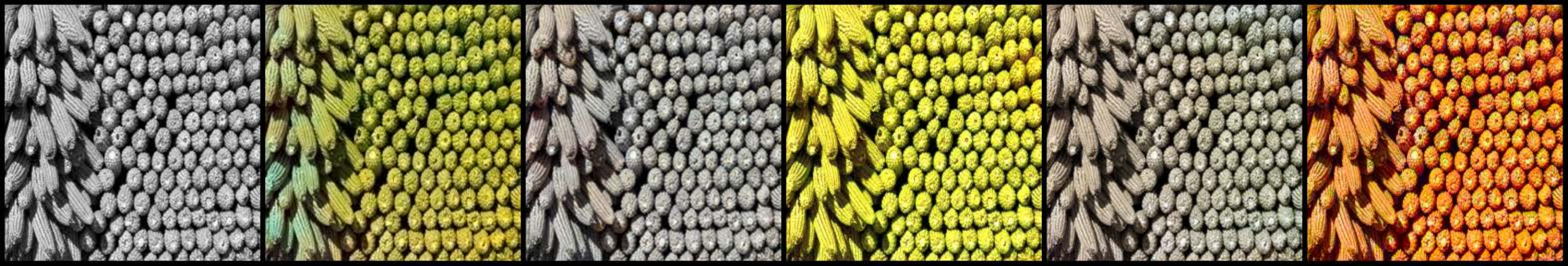}
	\caption{User preferences for different methods. \\From column 2-6: $7\%$(CIC) : $0\%$(ChromaGAN) : $20\%$(DeOldify) : $0\%$(InstColor) : $73\%$(GCP-Colorization).}
\end{figure*}

\begin{figure*}
	\centering
	\captionsetup{justification=centering}
	\includegraphics[width=\linewidth]{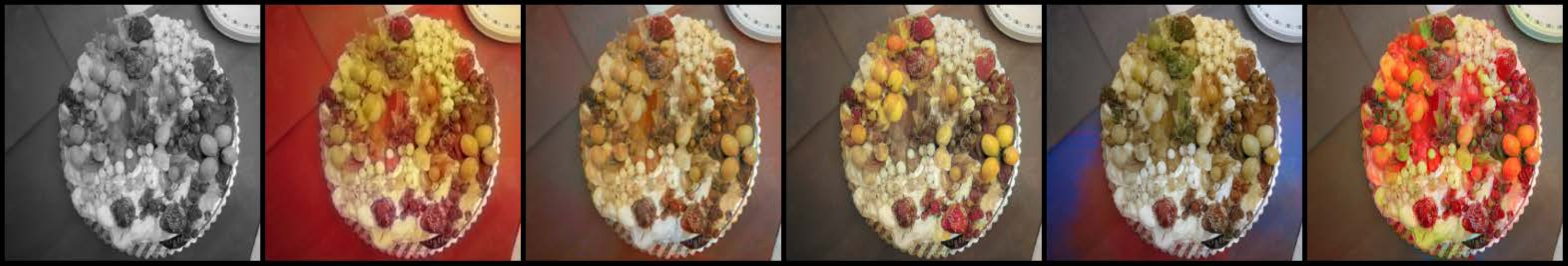}
	\caption{User preferences for different methods. \\From column 2-6: $3\%$(CIC) : $0\%$(ChromaGAN) : $23\%$(DeOldify) : $13\%$(InstColor) : $60\%$(GCP-Colorization).}
\end{figure*}

\begin{figure*}
	\centering
	\captionsetup{justification=centering}
	\includegraphics[width=\linewidth]{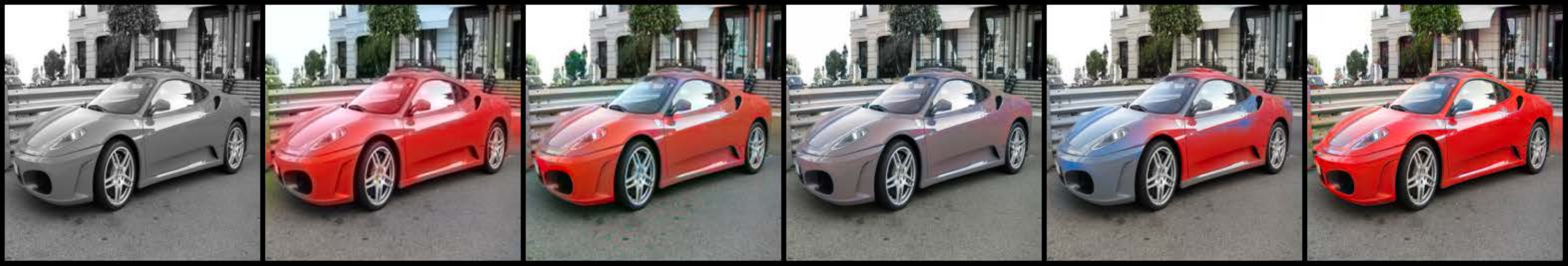}
	\caption{User preferences for different methods. \\From column 2-6: $0\%$(CIC) : $13\%$(ChromaGAN) : $13\%$(DeOldify) : $0\%$(InstColor) : $73\%$(GCP-Colorization).}
\end{figure*}

\begin{figure*}
	\centering
	\captionsetup{justification=centering}
	\includegraphics[width=\linewidth]{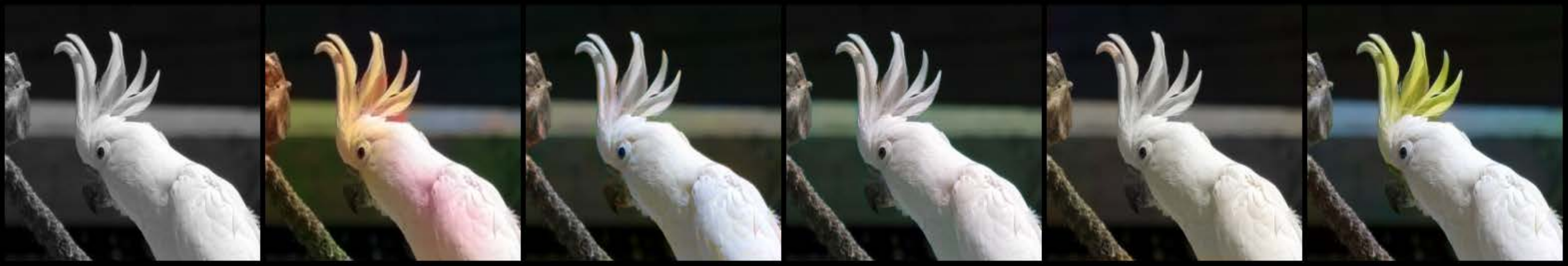}
	\caption{User preferences for different methods. \\From column 2-6: $13\%$(CIC) : $0\%$(ChromaGAN) : $10\%$(DeOldify) : $13\%$(InstColor) : $63\%$(GCP-Colorization).}
\end{figure*}

\begin{figure*}
	\centering
	\captionsetup{justification=centering}
	\includegraphics[width=\linewidth]{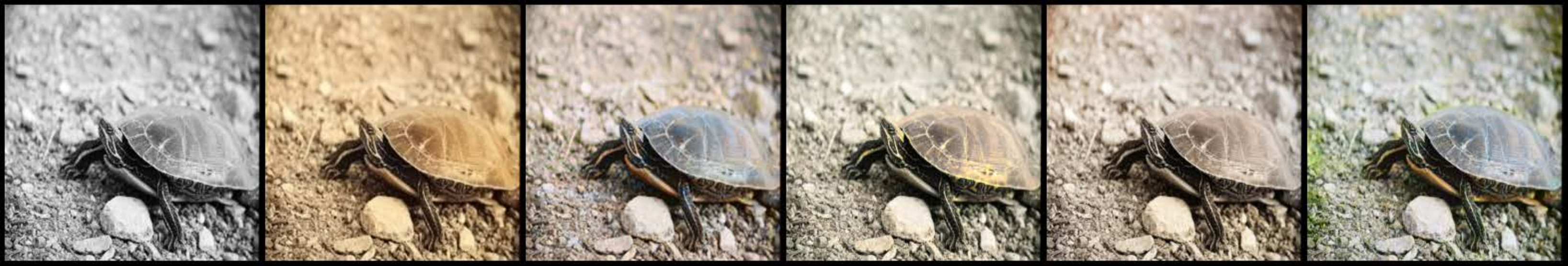}
	\caption{User preferences for different methods. \\From column 2-6: $0\%$(CIC) : $30\%$(ChromaGAN) : $0\%$(DeOldify) : $0\%$(InstColor) : $70\%$(GCP-Colorization).}
\end{figure*}

\begin{figure*}
	\centering
	\captionsetup{justification=centering}
	\includegraphics[width=\linewidth]{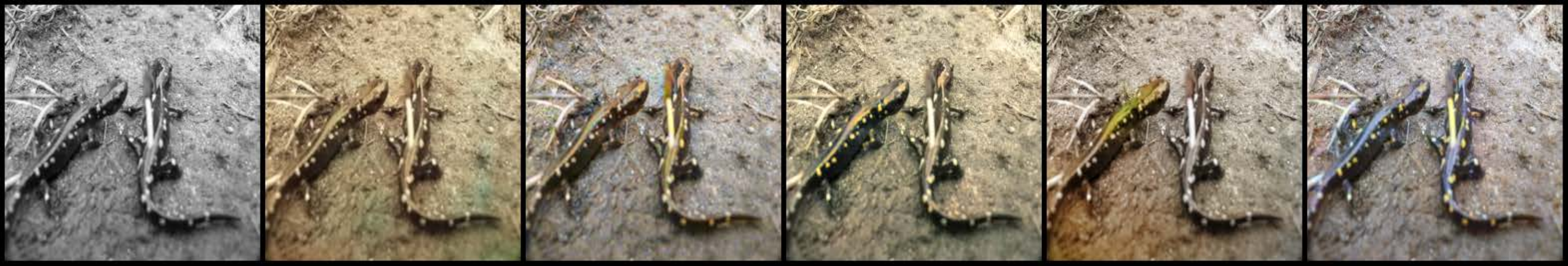}
	\caption{User preferences for different methods. \\From column 2-6: $0\%$(CIC) : $10\%$(ChromaGAN) : $7\%$(DeOldify) : $0\%$(InstColor) : $47\%$(GCP-Colorization).}
\end{figure*}

\begin{figure*}
	\centering
	\captionsetup{justification=centering}
	\includegraphics[width=\linewidth]{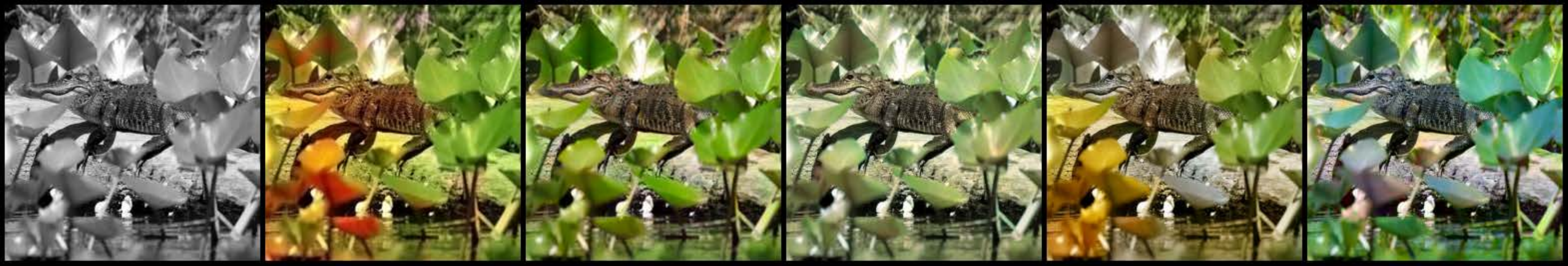}
	\caption{User preferences for different methods. \\From column 2-6: $3\%$(CIC) : $43\%$(ChromaGAN) : $0\%$(DeOldify) : $7\%$(InstColor) : $83\%$(GCP-Colorization).}
\end{figure*}

\begin{figure*}
	\centering
	\captionsetup{justification=centering}
	\includegraphics[width=\linewidth]{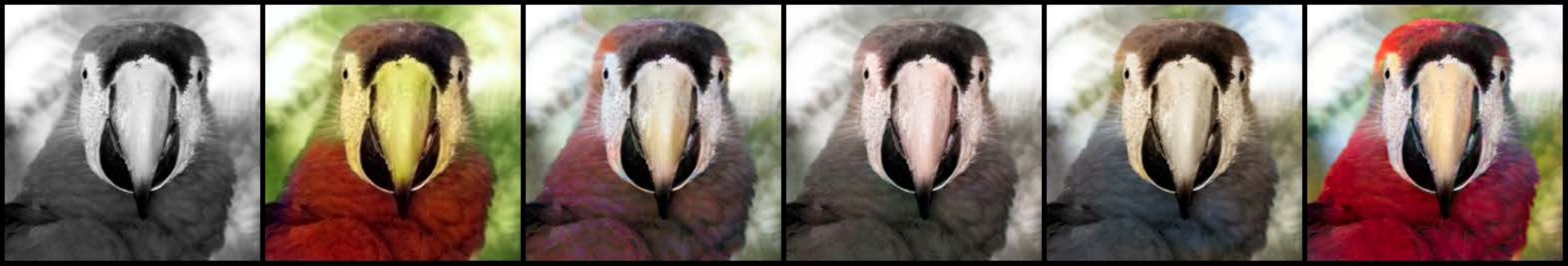}
	\caption{User preferences for different methods. \\From column 2-6: $7\%$(CIC) : $0\%$(ChromaGAN) : $3\%$(DeOldify) : $10\%$(InstColor) : $80\%$(GCP-Colorization).}
\end{figure*}

\begin{figure*}
	\centering
	\captionsetup{justification=centering}
	\includegraphics[width=\linewidth]{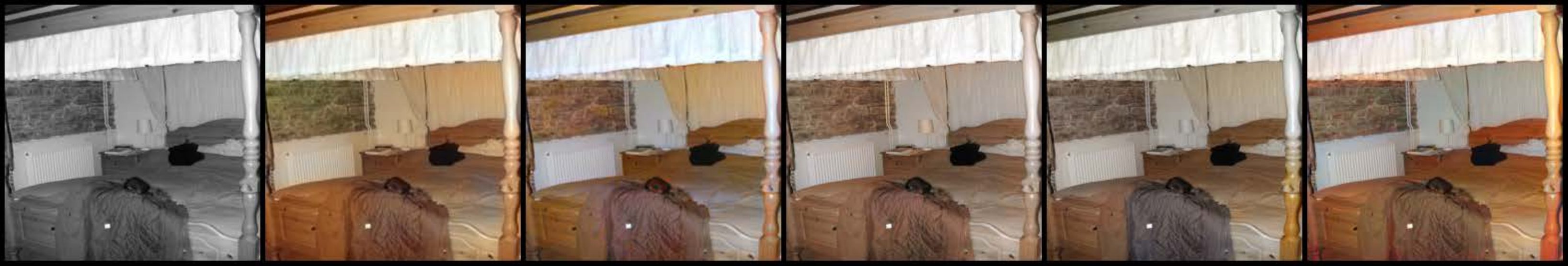}
	\caption{User preferences for different methods. \\From column 2-6: $10\%$(CIC) : $3\%$(ChromaGAN) : $0\%$(DeOldify) : $0\%$(InstColor) : $87\%$(GCP-Colorization).}
\end{figure*}

\begin{figure*}
	\centering
	\captionsetup{justification=centering}
	\includegraphics[width=\linewidth]{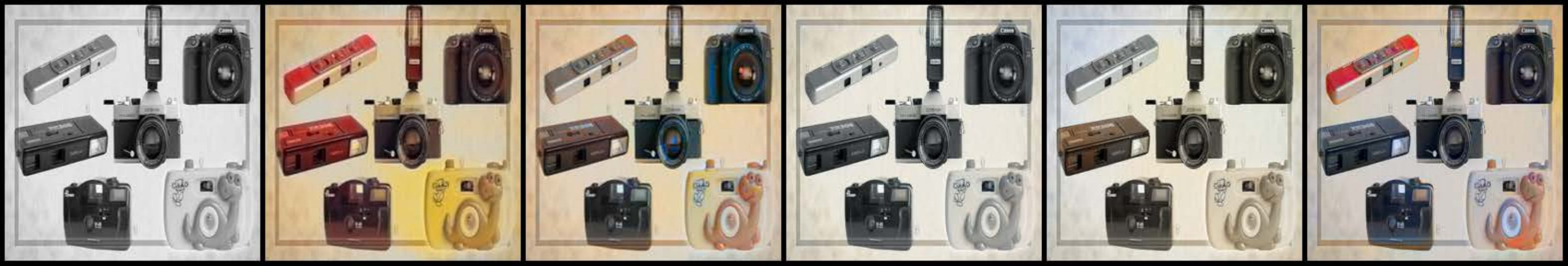}
	\caption{User preferences for different methods. \\From column 2-6: $10\%$(CIC) : $23\%$(ChromaGAN) : $3\%$(DeOldify) : $0\%$(InstColor) : $63\%$(GCP-Colorization).}
\end{figure*}

\begin{figure*}
	\centering
	\captionsetup{justification=centering}
	\includegraphics[width=\linewidth]{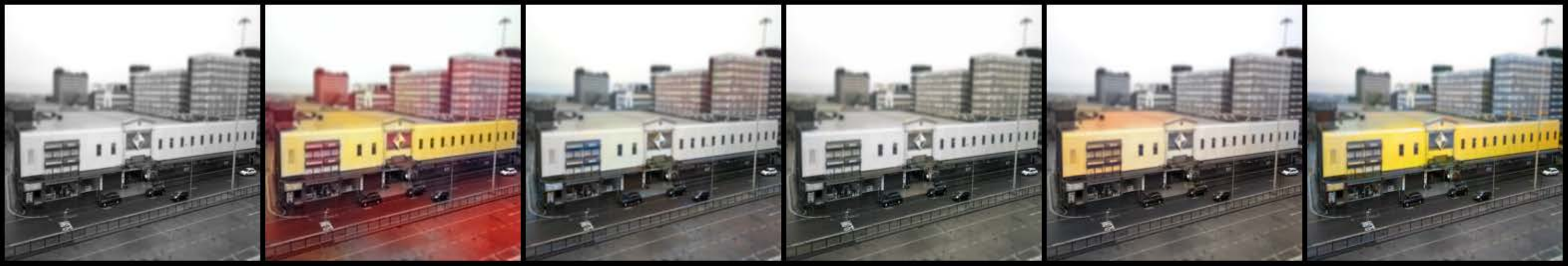}
	\caption{User preferences for different methods. \\From column 2-6: $17\%$(CIC) : $7\%$(ChromaGAN) : $3\%$(DeOldify) : $3\%$(InstColor) : $70\%$(GCP-Colorization).}
\end{figure*}

\end{document}